\documentclass[letterpaper]{article} 
\usepackage[preprint]{aaai2027} 
\usepackage[hyphens]{url}  
\usepackage{graphicx} 
\usepackage{natbib}  
\usepackage{caption} 
\usepackage{amsmath,amssymb,amsfonts,bm}
\usepackage{booktabs}
\usepackage{multirow}
\usepackage{array}

\newcommand{\proxyreg}{\textsc{ProxyReg}}
\newcommand{\fullmethod}{\textsc{Evi-VN}}
\newcommand{\R}{\mathbb{R}}
\newcommand{\V}{\mathcal{V}}
\newcommand{\E}{\mathcal{E}}
\newcommand{\T}{\mathcal{T}}
\newcommand{\Ddev}{\mathcal{D}_{\mathrm{dev}}}
\newcommand{\Dtest}{\mathcal{D}_{\mathrm{test}}}
\newcommand{\OOF}{\mathrm{OOF}}
\newcommand{\softmax}{\operatorname{softmax}}
\newcommand{\sigmoid}{\sigma}
\newcommand{\logit}{\operatorname{logit}}
\newcommand{\clip}{\operatorname{clip}}
\newcommand{\stopgrad}{\operatorname{sg}}

\title{Evi-VN: Hard Region Guided Virtual Node Evidence Injection for GNN-Based Fraud Detection}
\author{Jiran Tao, Yifan Wu, Binyan Jiang}
\affiliations{
The Hong Kong Polytechnic University\\
\{22117758r, yat-fan.wu\}@connect.polyu.hk, by.jiang@polyu.edu.hk
}

\begin{document}
\pagestyle{plain}
	
	\maketitle
	
	\begin{abstract}
		Online platforms contain growing numbers of bots, deceptive reviewers, and scam accounts that imitate legitimate users. Such camouflage blurs graph neighborhoods and behavioral attributes, making it difficult for graph neural networks (GNNs) to distinguish both well-disguised fraudsters and legitimate users. Across diverse GNNs, we observe overlapping errors on a shared hard region, suggesting the presence of latent fraud evidence that graph topologies and standard features fail to capture. Fraud-specific GNNs can mitigate particular graph pathologies, yet they still make limited use of heterogeneous evidence such as structured records, text, images, and audio; uniform multimodal fusion may also disturb nodes already handled reliably by the graph. We propose \fullmethod{} to learn and correct these shared blind spots rather than build another fraud detector. To our knowledge, \fullmethod{} is the first graph fraud detection framework to use feature isolated evidence chains to correct hard regions shared across GNNs. Its evidence chains connect behavior, content, and context across structured, textual, visual, and acoustic sources, helping expose camouflage that graph neighborhoods may miss. Crucially, \fullmethod{} selectively applies this evidence only to likely hard samples via virtual class nodes, preserving both the reliable predictions and the input design of existing GNNs. Shared hard regions also let \fullmethod{} enhance generic, fraud-specific, and unseen GNNs even with imperfect evidence models. Experiments across bot, fake-review, refund-evidence, and telecom-fraud tasks validate these advantages.
	\end{abstract}

	\section{Introduction}
	
	Graph-based fraud detection methods use relations among accounts, reviews, products, claims, and activities to capture coordinated behavior \cite{motie2024financial,cheng2024gnnfraudreview}. However, camouflaged fraudsters can imitate benign connections and attributes, while atypical benign users may appear suspicious, leaving some nodes difficult for structurally driven models. Fraud datasets also contain complementary behavioral records, text, images, and audio that standard GNN inputs may not fully exploit. Uniformly fusing such evidence into every node may also disturb predictions already handled reliably by the graph. Across diverse backbones, we observe substantially overlapping out-of-fold (OOF) errors and uncertain predictions, which we define as a GNN-common hard region. 
    Consequently, the fundamental question is not how to fuse every available feature into every node, but how to identify these intrinsically hard nodes and leverage information from other modalities to resolve these systematic failures of GNNs.

	We propose \fullmethod{}, a novel plug-and-play enhancement framework for
	existing GNN-based fraud detectors. It organizes complementary behavioral, content, and contextual cues into feature isolated evidence chains and applies them only to likely hard nodes, avoiding perturbation of confident GNN predictions. This framework consists of three key components: a router that identifies likely hard nodes, a replaceable evidence processor that produces a calibrated class prior, and a pair of virtual class nodes, with only the prior-selected class node injecting a source-only message into each routed hard node to form the augmented graph. The evidence processor communicates with the graph learner only through a fixed scalar prior interface; evidence
	features are never appended to the original node feature matrix, and the
	backbone sampling and aggregation mechanisms remain unchanged. This
	design allows \fullmethod{} to augment diverse generic and fraud-specific
	GNN backbones without redesigning their original input schema.

	We evaluate \fullmethod{} on bot detection, fake-review detection,
	refund-evidence fraud, and telecom-fraud tasks with structured, textual,
	visual, and acoustic evidence. Across these settings, \fullmethod{}
	consistently improves the matched backbone-only baselines, and the gains
	extend to both generic and fraud-specific GNNs, including held-out
	architectures not used for hard-region discovery. Additional experiments
	show that the framework remains effective under alternative router and
	prior-generator families.
	
	Our contributions are (1) we identify and quantify
	GNN-common hard regions using multi-backbone out-of-fold predictions;
	(2) we propose a plug-and-play, feature-isolated enhancement framework
	that routes likely hard nodes and injects calibrated evidence through
	virtual class nodes; and (3) we demonstrate consistent gains across
	multiple fraud domains, evidence modalities, and both seen and held-out
	GNN backbones.
	
	\section{Related Work}
	
	Graph-based fraud detection methods capture coordinated fraudulent behavior that is invisible in isolated records. CARE-GNN mitigates camouflage through relation-aware neighbor selection \cite{dou2020enhancing}, PC-GNN addresses class imbalance through label-balanced sampling \cite{liu2021pcgnn}, and H$^{2}$-FDetector separately models homophilic and heterophilic connections \cite{shi2022h2fdet}. Social-bot benchmarks such as TwiBot-20, TwiBot-22, Cresci-2015, and MGTAB emphasize account relations and behavioral attributes \cite{feng2021twibot20,feng2022twibot22,cresci2015fame,shi2023mgtab}, while collective review detection similarly exploits reviewer--product structure and metadata \cite{rayana2015collective}. These methods improve graph learning under specific structural pathologies but leave non-graph evidence largely unused.
	
	In particular, recent fraud datasets extend beyond graph relations and tabular attributes to include heterogeneous evidence such as text, images, and behavioral records. FraudBench introduces image--text evidence for fraudulent refund claims \cite{yan2026fraudbench}, building on textual and behavioral information already present in social-bot and review benchmarks \cite{feng2021twibot20,rayana2015collective}. Stage-based fusion taxonomies traditionally distinguish early, late, and hybrid fusion \cite{jiao2024comprehensive}, though recent surveys argue that these stage-based categories no longer capture modern deep multimodal architectures \cite{zhao2024deep}. For the GNN setting, the two endpoints of this spectrum already expose the design tension: early fusion couples GNN inputs to modality encoders, whereas late fusion preserves the backbone but cannot influence message passing \cite{jiao2024comprehensive}. To resolve this, \fullmethod{} instead injects evidence structurally via virtual nodes. By selectively targeting on nodes that   GNNs struggle to classify, our approach allows multimodal evidence to effectively guide message passing, improving detection accuracy on difficult cases without disrupting already reliable predictions.
	
	Virtual nodes offer a structural channel that can avoid this dichotomy, but they are typically studied as global structural augmentations. A message-passing neural network with a virtual node can approximate graph-transformer attention \cite{cai2023connection}, yet the standard virtual node assigns uniform node importance \cite{southern2025virtualnodes}, prompting learnable rewiring \cite{qian2024probabilistic} or clustered multiple virtual nodes \cite{hwang2022analysis}. These designs still treat the virtual node as an unsupervised aggregator of node features. LLM-guided virtual-node edge prediction also converts propagation-path semantics into graph structure for rumor detection \cite{tao2026llmrumor}. Less attention has been paid to virtual nodes as a conditioned channel that injects sample-specific external evidence only into a routed subset of real nodes. \fullmethod{} closes this gap by using class-specific virtual nodes whose fixed prototype features and source-only edges deliver a calibrated prior only at routed hard nodes.
	
	\section{Method}
	We introduce \fullmethod{} to incorporate external evidence into graph representation learning only for nodes estimated to be insufficiently resolved by the graph. A router identifies graph-hard nodes, a replaceable evidence processor estimates calibrated class priors, and two class-specific virtual nodes inject these priors while being regularized as bounded class proxies.
	
    Let $G=(\V,\E)$ be a graph with $n=|\V|$ real nodes. Each node
    $i\in\V$ has graph features $\mathbf{x}_i^{G}\in\R^{d_G}$ and
    a binary label $y_i\in\{0,1\}$. Training, development, and test splits
    are denoted by $\T$, $\Ddev$, and $\Dtest$. Each GNN fit median-imputes and standardizes graph-side features
    using only its fitting nodes: $\T\setminus\T_k$ for an OOF model and
    all of $\T$ for a full refit. The fitted transformation is then applied
    unchanged to held-out, development, and test nodes. Throughout,
    $\mathbf{x}_i^{G}$ denotes the corresponding processed vector and
    $\mathbf{X}^{G}=[\mathbf{x}_1^{G},\ldots,\mathbf{x}_n^{G}]^\top$.
	
	The graph learner takes $(G,\mathbf{X}^{G})$ as input. External evidence $\mathbf{s}_i^E$ is used exclusively for routing and prior estimation. It is never concatenated with GNN inputs or passed to the final classifier. Its construction is dataset-specific and may involve structured, textual, visual, or acoustic modalities; Supplementary Section~S4.2 gives the extraction details and leakage controls.
	
	This framework consists of three key components: a router that identifies likely hard nodes, a replaceable evidence processor that produces a calibrated class prior, and a pair of virtual class nodes.
    For each routed node, the prior selects one class virtual node and creates a source-only edge, where the virtual node is the message source and the real node is the target, with no reverse edge. These directed prior edges form the augmented graph.
	\begin{equation*}
		\begin{aligned}
			f_\theta &: (G,\mathbf{X}^{G}) \mapsto (\mathbf{Z}^{G},\mathbf{H}^{G}), \\
			g_\psi &: \mathbf{s}_i^{E} \mapsto r_i \in [0,1], \\
			h_\phi   &: \mathbf{s}_i^{E} \mapsto q_i \in [0,1].
		\end{aligned}
	\end{equation*}
	Here $f_\theta$ is the target graph learner, which outputs the raw prediction scores $\mathbf{Z}^{G}\in \R^{n\times 2}$ and the corresponding node embeddings $\mathbf{H}^{G}$.
    $r_i$ estimates graph hardness, and $q_i$ is an evidence-derived class prior. Non-routed nodes keep a neutral prior; routed nodes use the prior from $h_\phi$. The evidence branch communicates with the graph learner only through the scalar pair $(r_i,q_i)$, so the modality-specific encoder can be replaced without altering the GNN feature schema.
	
	During training, out-of-fold predictions from multiple discovery
	backbones provide supervision for the router. The learned routing
	scores and calibrated priors determine the source-only edges from two
	class-specific virtual nodes to the real nodes. The augmented graph
	is then optimized with the task loss and our proposed proxy regularization (\proxyreg{}). The discovery
	backbones serve only to construct routing targets and are discarded
	afterwards; inference requires only the target graph learner, the
	router, and the prior generator for routed nodes.
	
	\begin{figure*}[t]
		\centering
		\includegraphics[width=0.99\textwidth]{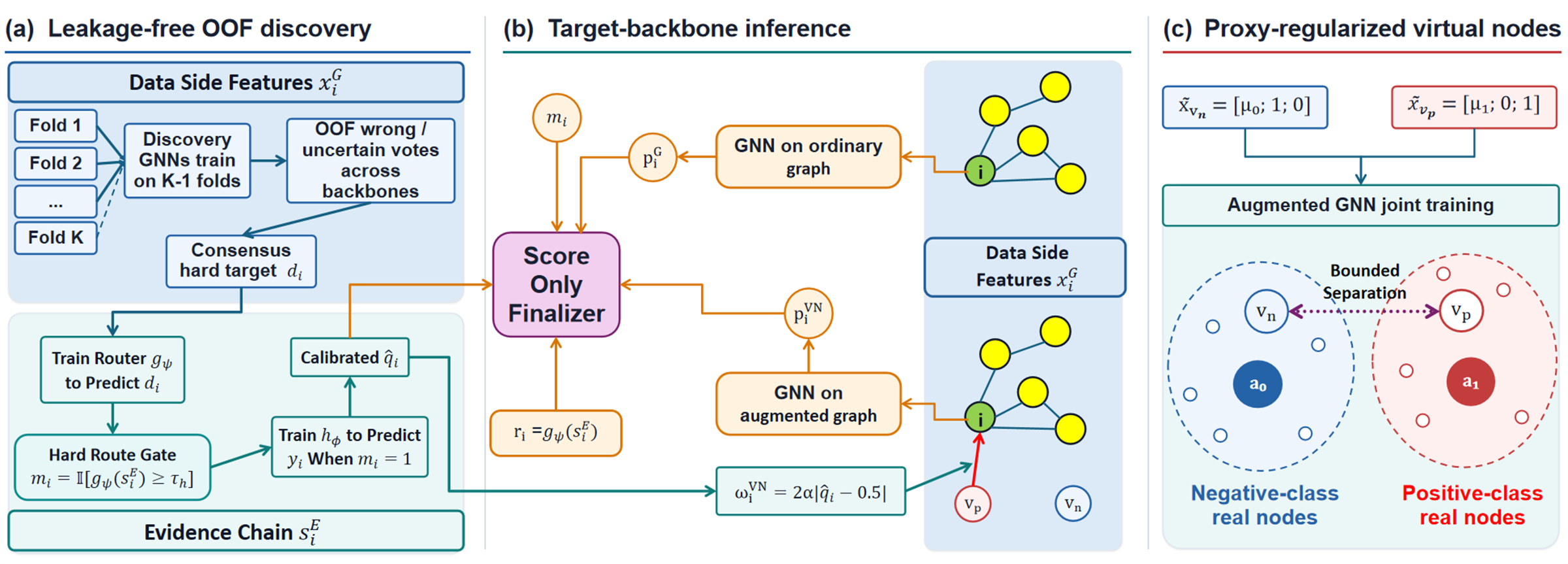}
		\caption{Overview of \fullmethod{}. (a) Discovery-GNN OOF errors define consensus-hard targets; the evidence router gates prior generation and calibration. (b) Original and augmented versions of one target backbone run in parallel. A calibrated prior selects and weights a source-only virtual edge, where $v_p$ and $v_n$ denote the positive- and negative-class virtual nodes, respectively. The finalizer combines fold-held-out graph, prior, and routing scalars. (c) Fixed class-prototype inputs become hidden virtual-node proxies under proxy alignment, bounded separation, and stop-gradient anchors ($a_0$ and $a_1$). Discovery backbones are training-only, and evidence-chain features never enter the GNN input.}
		\label{fig:evivn-framework}
	\end{figure*}
	
	\subsection{Hard Gated Leakage Free Evidence Priors}
	
	The router and the prior generator form a coupled, hard gated prior pipeline. The multi-backbone hard region diagnosis is used to obtain $g_\psi$; it is not the direct training target of the prior generator. Let $\mathcal{B}$ be the set of discovery backbones and let $p_{ib}^{G,\OOF}$ be the out-of-fold positive class probability of backbone $b$ on training node $i$. Define $\hat y_{ib}^{\OOF}=\mathbb{I}[p_{ib}^{G,\OOF}\ge 0.5]$, $e_{ib}=\mathbb{I}[\hat y_{ib}^{\OOF}\ne y_i]$, and $u_{ib}=\mathbb{I}[|p_{ib}^{G,\OOF}-0.5|\le\delta]$. A training node is marked consensus hard when enough discovery backbones are wrong or uncertain:
	\begin{equation*}
		\begin{aligned}
			d_i=\mathbb{I}\!\Big[&
			\textstyle\sum_{b\in\mathcal{B}} e_{ib}\ge\kappa_{\mathrm{err}}\\[-2pt]
			&\vee\ \textstyle\sum_{b\in\mathcal{B}}
			\max(e_{ib},u_{ib})\ge\kappa_{\mathrm{hard}}
			\Big],\quad i\in\T .
		\end{aligned}
	\end{equation*}
	This uses only training OOF predictions and labels; setting $u_{ib}=0$ disables uncertainty votes.
	
	The hard/easy router $g_\psi$ is then trained to predict $d_i$ from the evidence chain feature vector $\mathbf{s}_i^{E}$. To keep the downstream prior leakage free on training nodes, router scores on $\T$ are also produced out of fold. For a stratified partition $\{\T_1,\ldots,\T_K\}$ of $\T$,
	\begin{equation*}
		r_i^{\OOF}=g_{\psi}^{(-k)}(\mathbf{s}_i^{E}),
		\qquad i\in\T_k,
	\end{equation*}
	where $g_{\psi}^{(-k)}$ is trained on $\T\setminus\T_k$. 
	
	A node is routed to the evidence chain prior generator only if
	\begin{equation*}
		m_i=\mathbb{I}[r_i\ge \tau_h]=1,
	\end{equation*}
	where $\tau_h$ is selected by development final-prediction accuracy over a fixed uniform grid, then fixed before test evaluation. Nodes with $m_i=0$ bypass the prior generator, receive $\hat q_i=0.5$, and have no virtual edge.

    The router need not solve fraud detection itself: it only estimates whether graph evidence is likely insufficient. Mathematically, the indicator $d_i$ flags nodes where GNN backbones are incorrect or highly uncertain, directly implying that the graph structure alone is insufficient for fraud detection.
    The router is trained to predict this target using strictly the external evidence chain $\mathbf{s}_i^{E}$ as input, outputting a routing score $r_i$. Thus $r_i$ acts as an independent, evidence-based estimate of graph failure. By thresholding this estimate, the indicator  
    $m_i$ acts as a gating mechanism, ensuring that structural intervention from the external evidence is injected only into nodes that actually need it.
    
    Nodes with \(m_i=1\) enter the second stage model, whose prior generator is trained on the routed subset to predict \(y_i\). This separates hard-region recognition (\(d_i\)) from (\(y_i\)) and allows the two stages to use different model families. Its training-node scores are cross-fitted:
	\begin{equation*}
		q_i^{\OOF}=h_{\phi}^{(-k)}(\mathbf{s}_i^{E}),
		\qquad i\in\T_k,\ m_i=1.
	\end{equation*}
	For development/test nodes, the prior generator predicts
	\begin{equation*}
		q_i^{\mathrm{raw}}=h_{\phi}^{(\T_h)}(\mathbf{s}_i^{E}),
		\qquad i\in\Ddev\cup\Dtest,\ m_i=1,
	\end{equation*}
	where $\T_h=\{i\in\T:m_i=1\}$. Thus the split-aware raw prior is
	\begin{equation*}
		q_i=
		\begin{cases}
			q_i^{\OOF}, & i\in\T,\ m_i=1,\\
			q_i^{\mathrm{raw}}, & i\in\Ddev\cup\Dtest,\ m_i=1,\\
			0.5, & m_i=0.
		\end{cases}
	\end{equation*}
	The prior is calibrated by a one-dimensional calibration function fitted only on training OOF predictions and labels:
	\begin{equation*}
		\hat q_i=
		\begin{cases}
			C(q_i), & m_i=1,\\
			0.5, & m_i=0,
		\end{cases}
		\qquad
		C(q)=\sigmoid(\alpha_C\logit(q)+\beta_C).
	\end{equation*}
	Only $\hat q_i$ enters graph injection and finalization. Component families are selected by downstream \fullmethod{} performance rather than standalone router/prior accuracy: a prior may be imperfect globally yet useful on routed graph failures. The evidence vector $\mathbf{s}_i^E$ is never appended to $\mathbf{X}^{G}$.
	
    \subsection{Virtual Node Construction and Graph Augmentation}

To convert the calibrated prior into graph messages, we introduce a
negative-class virtual node $v_n$ and a positive class virtual node
$v_p$. For $c\in\{0,1\}$, define
$\T^{(c)}=\{i\in\T:y_i=c\}$ as the class-$c$ subset of the full
training split. In the $k$th OOF fit, every prototype and loss below instead
uses
$(\T\setminus\T_k)^{(c)}
=\{i\in\T\setminus\T_k:y_i=c\}$,
so held-out nodes do not enter feature construction. For the full
refit, let
$\bm{\mu}_c=|\T^{(c)}|^{-1}
\sum_{i\in\T^{(c)}}\mathbf{x}_i^{G}$.
The nodes inputs are
\[
\tilde{\mathbf{x}}_i^{G}=[\mathbf{x}_i^{G};0;0],\quad
\tilde{\mathbf{x}}_{v_n}^{G}=[\bm{\mu}_0;1;0],\quad
\tilde{\mathbf{x}}_{v_p}^{G}=[\bm{\mu}_1;0;1].
\]
Let $\tilde{\mathbf{X}}^{G}$ stack these inputs over
$\V^+=\V\cup\{v_n,v_p\}$. The class means provide fixed input
prototypes rather than free embeddings; their hidden representations
are learned through the shared augmented GNN parameters.

Before message passing, $\hat q_i$ selects the virtual node source and
edge strength. For thresholds $\tau_1>0.5$ and $\tau_0<0.5$, define
	\begin{equation*}
		\xi_i = 2|\hat q_i-0.5| \in [0,1].
	\end{equation*}
The source index is
	\begin{equation*}
		\chi_i=
		\begin{cases}
			p, & m_i=1 \text{ and } \hat q_i\ge \tau_1,\\
			n, & m_i=1 \text{ and } \hat q_i\le \tau_0,\\
			\varnothing, & \text{otherwise}.
		\end{cases}
	\end{equation*}
    Here $p$ represents the positive virtual node $v_p$, $n$
represents the negative virtual node $v_n$, and $\varnothing$ indicates no
injected edge.
The calibrated prior thus augments the original graph by constructing
the source-only edge set $\E_{\mathrm{VN}}$. Each prior-induced edge
has weight $w_i^{\mathrm{VN}}=\alpha\xi_i$; layers without scalar edge
weights represent its strength using repeated edges. Combining these
prior-induced edges with the original graph gives
$G^+=(\V^+,\E^+,\tilde{\mathbf{X}}^{G})$, where
$\E^+=\E\cup\E_{\mathrm{VN}}$. The target backbone then operates on the augmented graph
$G^+$ to output the augmented prediction logit $\mathbf{z}_i^{\mathrm{VN}}\in \R^2$ for node $i$, yielding
$p_i^{\mathrm{VN}}
=\softmax(\mathbf{z}_i^{\mathrm{VN}})_1$. Here, the subscript $1$ selects the component associated with class label $y_i=1$, which denotes the fraud class.
	
This construction prevents real node messages from rewriting the class prototypes and then broadcasting graph noise globally. The confidence weighting $w_i^{\mathrm{VN}}$ further quantifies the prior strength, ensuring that uncertain evidence produces weak or no structural intervention.
	
	\subsection{Real Node Classification Loss}
	
	Virtual nodes are unlabeled and excluded from all data masks. Supervision uses real training nodes only:
	\begin{equation*}
		\begin{split}
			\mathcal{L}_{\mathrm{CE}}
			&=-\frac{1}{|\T|}
			\sum_{i\in\T}\omega_{y_i}
			\log\softmax(\mathbf{z}_i^{\mathrm{VN}})_{y_i},\\
			\omega_c
            &=\frac{|\T|}{2\max(|\T^{(c)}|,1)} .
		\end{split}
	\end{equation*}
	
	\subsection{Virtual Node Proxy Regularization}

To ensure the virtual nodes effectively represent their respective classes, we introduce \proxyreg{} to encourage the virtual nodes to serve as class proxies in the hidden space. Notably, their input features are set to be the fixed class-feature prototypes introduced in the virtual node construction section, rather than freely trainable node embeddings. During the training of the augmented GNN (i.e., the GNN operating on the augmented graph), their hidden representations are recomputed through the shared message passing layers and evolve with the backbone parameters. They are not directly assigned the mean hidden representation of either class.

	Without this constraint, virtual nodes can act as ordinary hubs whose semantics are only implied by their initial features. Proxy classification aligns their learned hidden representations with real classes, while bounded separation and anchoring avoid unrestricted prototype drift or collapse.
	
	Let $\mathbf{u}_j=\mathbf{h}_j/(\|\mathbf{h}_j\|_2+\epsilon)$ denote the normalized augmented GNN hidden representation of either a real or virtual node $j$, and let $\pi(0)=n,\pi(1)=p$. The similarity of real node $i$ to class proxy $c$ is $s_{i,c}=\mathbf{u}_i^\top\mathbf{u}_{v_{\pi(c)}}/\tau_p$.
	The proxy classification loss is
	\begin{equation*}
		\mathcal{L}_{\mathrm{proxy}}
		=-\frac{1}{|\T|}\sum_{i\in\T}\omega_{y_i}
		\log\frac{\exp(s_{i,y_i})}{\exp(s_{i,0})+\exp(s_{i,1})}.
	\end{equation*}
	
	We use bounded separation
\begin{equation*}
\mathcal{L}_{\mathrm{sep}}
=
\left[
\mathbf{u}_{v_n}^{\top}\mathbf{u}_{v_p}
-\gamma_{\mathrm{sep}}
\right]_+^2,
\end{equation*}
where $\gamma_{\mathrm{sep}}\in[-1,1]$ is the maximum allowed cosine
similarity between the two virtual node proxies. The loss is zero when
their similarity is at most $\gamma_{\mathrm{sep}}$ and penalizes only
excess similarity, preventing proxy collapse without forcing
unbounded separation.

	For each class, let $\bar{\mathbf{u}}_c
=|\T^{(c)}|^{-1}\sum_{i\in\T^{(c)}}\mathbf{u}_i .$ be its current real-node center and define the corresponding anchor as $\mathbf{a}_c=\stopgrad(\bar{\mathbf{u}}_c)$. The anchor loss is
\begin{equation*}
\begin{split}
\mathcal{L}_{\mathrm{anchor}}
=\sum_{c\in\{0,1\}}
\Big[
\big\|\mathbf{u}_{v_{\pi(c)}}-\mathbf{a}_c\big\|_2
-\rho_{\mathrm{anc}}
\Big]_+^2 .
\end{split}
\end{equation*}
Here $\rho_{\mathrm{anc}}\in[0,2]$ is the allowed radius around the
stop-gradient class anchor. The anchor
$\mathbf{a}_c=\stopgrad(\bar{\mathbf{u}}_c)$ is recomputed from the
current real node representations but receives no gradient. The loss
is zero while the corresponding virtual proxy remains within
$\rho_{\mathrm{anc}}$ of its class anchor and penalizes only the
excess distance. It therefore limits proxy drift without assigning
the class center directly to the virtual node or moving real nodes
through the anchor.
	
	\subsection{Proxy Regularized GNN Training Objective}
	
	The augmented GNN is trained with the objective in Eq.~\eqref{eq:total-loss}:
	\begin{equation}
		\label{eq:total-loss}
		\begin{split}
			\mathcal{L}
			&=\mathcal{L}_{\mathrm{CE}}
			+\eta_t\left(
			\lambda_{\mathrm{proxy}}\mathcal{L}_{\mathrm{proxy}}
			+\lambda_{\mathrm{sep}}\mathcal{L}_{\mathrm{sep}}\right.\\
			&\hspace{2.6em}\left.
			+\lambda_{\mathrm{anchor}}\mathcal{L}_{\mathrm{anchor}}
			\right),
		\end{split}
	\end{equation}
	where $\eta_t=\min(1,t/T_{\mathrm{warm}})$. Hyperparameters and component checks are reported in Supplementary Section S3.
	
	\subsection{Score Only Finalizer}
	\label{sec:score-only-finalizer}
	
	Let $p_i^{G}$ and $p_i^{\mathrm{VN}}$ be the original- and augmented-graph probabilities. The finalizer uses the five scalars in Eq.~\eqref{eq:score-vector}:
	\begin{equation}
		\label{eq:score-vector}
		\begin{aligned}
			\mathbf{s}_i^{F}=
			[\,
			\ell_i^{\mathrm{VN}};
			\ell_i^{q};
			\ell_i^{G};
			r_i;
			m_i\,],
		\end{aligned}
	\end{equation}
	where
	\begin{align*}
		\ell_i^{\mathrm{VN}}
		&= \log\frac{p_i^{\mathrm{VN}}}{1-p_i^{\mathrm{VN}}},\\
		\ell_i^{G}
		&= \log\frac{p_i^{G}}{1-p_i^{G}},\\
		\ell_i^{q}
		&= \clip\!\left(
		\log\frac{\hat q_i}{1-\hat q_i},-B,B
		\right).
	\end{align*}
	For easy nodes, $m_i=0$, $\hat q_i=0.5$, and $\ell_i^q=0$. Every training component is cross-fitted. For $i\in\T_k$, $p_i^{G}$ and $p_i^{\mathrm{VN}}$ come from original and augmented GNNs trained only on $\T\setminus\T_k$; the matching $r_i$ and $\hat q_i$ are also OOF. Development/test components come from models refit on all of $\T$. Thus the finalizer sees fold-held-out base-score distributions during training, rather than the overconfident predictions of GNNs evaluated on their own training nodes. This is stacking control, not merely label isolation: without it, a leakage-free test set could still face a train--test confidence shift at the finalizer.
	
	The final probability is produced by the development-selected score-only classifier in Eq.~\eqref{eq:score-only-finalizer}:
	\begin{equation}
		\label{eq:score-only-finalizer}
		\hat p_i^{\mathrm{final}}=\rho_\omega(\mathbf{s}_i^{F}),
	\end{equation}
	where $\rho_\omega$ is a development-selected lightweight classifier trained with class-weighted supervision. Restricting it to scalar scores keeps the multimodal claim attributable to the prior and virtual-edge path rather than a second hidden multimodal classifier. It never consumes $\mathbf{s}_i^E$ or $\mathbf{h}_i^{\mathrm{VN}}$; all configurations and thresholds are fixed before test evaluation.
	
	\section{Experimental Setup}
	\label{sec:experimental-setup}
	
	\subsection{Datasets}
	
	We evaluate six public benchmarks spanning social bots, deceptive reviews, refund evidence, and telecom fraud: MGTAB \cite{shi2023mgtab}, TwiBot-20 \cite{feng2021twibot20}, Cresci-2015 \cite{cresci2015fame}, YelpZip \cite{rayana2015collective}, FraudBench \cite{yan2026fraudbench}, and TeleAntiFraud-28k \cite{ma2025teleantifraud}. MGTAB is the primary setting, with 10,199 labeled accounts and a 7,139/1,530/1,530 train/dev/test split; the others test transfer across structured, textual, visual, and acoustic evidence.
	
	All six datasets are publicly accessible and downloadable from their cited releases, so we do not upload or redistribute raw copies. Supplementary Section S4.2 reports dataset statistics, splits, graph construction, retained evidence, and excluded shortcut fields; Sections S5--S7 give the extended YelpZip, FraudBench, and TeleAntiFraud protocols.

	\subsection{Backbones and Prior Generators}
	
	MGTAB uses 13 discovery backbones: MLP, GCN \cite{kipf2017gcn}, GAT \cite{velickovic2018gat}, GraphSAGE \cite{hamilton2017graphsage}, GCNII \cite{gcnii}, APPNP \cite{klicpera2019appnp}, GPRGNN \cite{gprgnn}, MixHop \cite{abu2019mixhop}, SIGN \cite{frasca2020sign}, ChebNet \cite{chebnet}, GraphConv \cite{graphconv}, TAGCN \cite{tagcn}, and GIN \cite{xu2019gin}. Its XGBoost \cite{chen2016xgboost} router and prior generator use OOF consensus labels and routed samples, respectively, with OOF Platt calibration \cite{platt1999probabilistic}. Development-only router selection is reported in Supplementary Table S14. Held-out tests include CARE-GNN \cite{dou2020enhancing} and PC-GNN \cite{liu2021pcgnn}; Supplementary Section S3.3 varies the discovery-set size.
	
	TwiBot-20 uses the same discovery set and an XGBoost prior; Cresci-2015 uses 13-backbone discovery, an XGBoost router, and logistic prior. Supplementary Section S1 reports their full results with YelpZip.
	
	\subsection{Baselines}
	
	Because \fullmethod{} augments rather than replaces a detector, our primary unit of comparison is a matched GNN backbone with identical graph inputs. This includes strong fraud-oriented backbones in the unseen-transfer study. The remaining controls hold the routed prior fixed and ask whether gains come from ensembling, output fusion, score learning, or structural injection; they are more diagnostic than comparing unrelated leaderboard numbers obtained with different features and splits.
	
	Accordingly, Table~\ref{tab:mgtab-main} measures average improvement over a broad discovery family, while Table~\ref{tab:mgtab-unseen-backbone} tests transfer to held-out architectures, including CARE-GNN and PC-GNN. This paired protocol complements rather than avoids SOTA comparison: it asks whether those detectors themselves become stronger under the same data split. A leaderboard comparison against methods using different relations, features, or label access would not isolate that contribution.
	
	The main paper compares \fullmethod{} with:
	\begin{itemize}
		\item \textbf{Backbone only}: the first-stage graph learner trained on the original graph.
		\item \textbf{GNN probability ensemble}: an output level ensemble of the 13 first stage GNN backbones.
		\item \textbf{Weighted probability fusion}: $(1-\lambda)p_i^G+\lambda\hat q_i$.
		\item \textbf{Weighted logit fusion}: $\sigmoid((1-\lambda)\logit(p_i^G)+\lambda\logit(\hat q_i))$.
	\end{itemize}
	Ensemble weights and all model configurations are selected on development data; prior-based baselines share $\hat q_i$.
	
	\subsection{Evaluation Metrics}
	
	We report Accuracy, positive class F1, and AUC. The positive class is bot for TwiBot-20, Cresci-2015, and MGTAB, fake/spam review for YelpZip, fake-damaged evidence for FraudBench, and fraudulent call for TeleAntiFraud-28k.
	All router, prior, calibration, route, fusion, and finalizer choices are fixed using training OOF or development data. Test labels are used only once for reporting. Training-node inputs to the finalizer are cross-fitted for every base score, whereas development/test scores come from models refit on the full training split. To support reproducibility and practical use, our submission includes the \fullmethod{} implementation and Evi-VN Studio, a configurable application for applying the framework to real world fraud detection tasks.

	\section{Results and Analysis}
	\label{sec:results}
	
	\subsection{MGTAB Main Results}
	
	Table~\ref{tab:mgtab-main} averages the 13 MGTAB discovery backbones under the configuration above. Reporting the mean paired backbone effect is central to our goal: \fullmethod{} should improve a family of graph learners rather than win through one favorable architecture.
	
\begin{table}[t]
	\centering
	\begin{tabular}{@{}lccc@{}}
			\toprule
			Method & Acc. & F1$_{+}$ & AUC \\
			\midrule
			Backbone only & 0.8606 & 0.7649 & 0.9207 \\
			GNN ensemble & 0.9072 & 0.8325 & 0.9660 \\
			Prob. fusion & 0.9171 & 0.8579 & 0.9692 \\
			Logit fusion & 0.9172 & 0.8580 & 0.9695 \\
			\fullmethod{} & \textbf{0.9283} & \textbf{0.8707} & \textbf{0.9769} \\
			\bottomrule
		\end{tabular}
		\caption{MGTAB test performance averaged over 13 GNN backbones. Metrics are Accuracy, positive class F1, and AUC.}
		\label{tab:mgtab-main}
	\end{table}
	
	\fullmethod{} improves mean Accuracy/F1/AUC over Backbone only by 6.77/10.58/5.61 percentage points (pp), and still exceeds logit fusion by 1.10/1.27/0.74 pp. The ensemble tests whether GNN diversity is sufficient, while probability/logit fusion expose the same routed prior only after message passing. Their remaining gap supports converting evidence into graph structure.

To verify that our gains are consistent across architectures, Table~\ref{tab:mgtab-paired-tests-main} matches each Evi-VN model directly against its corresponding baseline. Specifically, for a given metric such as AUC, we calculate the exact performance difference between Evi-VN and the baseline for each of the 13 individual GNN architectures. We then apply bootstrapping to these 13 difference scores to construct 95\% confidence intervals, and use a one-sided sign-flip test to assess statistical significance. Notably, these tests confirm that \fullmethod{} significantly outperforms all baselines. This consistent superiority proves that selectively injecting evidence into the graph structure via virtual nodes is fundamentally more powerful than standard post-hoc score blending, regardless of the underlying GNN architecture. Supplementary Section S1.2 reports additional baseline comparisons on the other three real datasets.
	
\begin{table}[t]
	\centering
	{\small
			\setlength{\tabcolsep}{1mm}
			\begin{tabular}{@{}llccc@{}}
				\toprule
				Control & Metric & $\Delta$ (pp) & 95\% CI (pp) & SF $p$ \\
				\midrule
				Logit fus. & Acc. & +1.10 & [0.36,1.90] & 0.0106 \\
				Logit fus. & F1$_+$ & +1.27 & [0.23,2.38] & 0.0216 \\
				Logit fus. & AUC & +0.74 & [0.05,1.64] & 0.0398 \\
				\addlinespace[1pt]
				Prob. fus. & Acc. & +1.11 & [0.33,1.96] & 0.0133 \\
				Prob. fus. & F1$_+$ & +1.27 & [0.18,2.45] & 0.0284 \\
				Prob. fus. & AUC & +0.77 & [0.08,1.68] & 0.0304 \\
				\addlinespace[1pt]
				Backbone & Acc. & +6.77 & [4.00,9.91] & $1.22{\times}10^{-4}$ \\
				Backbone & F1$_+$ & +10.58 & [6.80,14.60] & $1.22{\times}10^{-4}$ \\
				Backbone & AUC & +5.61 & [3.05,8.56] & $1.22{\times}10^{-4}$ \\
				\bottomrule
			\end{tabular}
		}
		\caption{MGTAB paired backbone tests. Differences are percentage points (pp). CI: bootstrap over backbones; SF: one-sided sign-flip test for \fullmethod{} superiority.}
		\label{tab:mgtab-paired-tests-main}
	\end{table}
	
	\subsection{Hard-node correction.} Table~\ref{tab:mgtab-hard-correction} evaluates target-excluded consensus-hard nodes: for each target, the other 12 backbones mark nodes hard when wrong or uncertain, preventing the evaluated model from selecting its own errors. On these hard nodes, \fullmethod{} raises Accuracy by 13.82 pp.

	\begin{table}[!t]
		\centering
		\setlength{\tabcolsep}{1mm}
		\begin{tabular}{@{}lccc@{}}
			\toprule
			Hard subset & Base & \fullmethod{} & $\Delta$ Acc.   \\
			\midrule
			LOO consensus & 52.97 & 66.79 & +13.82   \\
			\bottomrule
		\end{tabular}
		\caption{MGTAB correction on target-excluded LOO consensus-hard nodes, averaged over 13 backbones. Accuracy is in percent.}
		\label{tab:mgtab-hard-correction}
	\end{table}

	\subsection{Router/prior replacement.} Table~\ref{tab:mgtab-router-prior-main} tests six plug-in pairings. Every pairing improves all metrics with modest variation, supporting component replaceability.

	\begin{table}[!t]
		\centering
		{\small
			\setlength{\tabcolsep}{1.5pt}
			\begin{tabular}{@{}llcc@{}}
				\toprule
				Router & Prior & Routed & $\Delta$ over backbone \\
				\midrule
				XGB  & XGB       & .8920/.8666/.9625 & +.0673/+.1056/+.0557 \\
				XGB  & LGBM      & .8929/.8682/.9622 & +.0660/+.1035/+.0545 \\
				XGB  & TabICL v2 & .8957/.8714/.9625 & +.0643/+.1015/+.0547 \\
				LGBM & XGB       & .8919/.8639/.9638 & +.0622/+.0973/+.0532 \\
				LGBM & LGBM      & .8955/.8679/.9630 & +.0635/+.1000/+.0537 \\
				RF   & XGB       & .8947/.8695/.9632 & +.0637/+.0997/+.0537 \\
				\bottomrule
			\end{tabular}
		}
		\caption{MGTAB router/prior replacement using XGBoost \cite{chen2016xgboost}, LightGBM \cite{ke2017lightgbm}, random forest, and TabICLv2 \cite{qu2026tabiclv2}. Each triplet reports Accuracy/F1/AUC; gains are over the matched 13-backbone Backbone only mean.}
		\label{tab:mgtab-router-prior-main}
	\end{table}
	
	\FloatBarrier
	\subsection{Cross-Dataset Results}
	
	Table~\ref{tab:cross-dataset-main} summarizes the additional tasks. TwiBot-20 and Cresci-2015 use structured evidence, YelpZip uses an LLM, FraudBench uses image--text evidence, and TeleAntiFraud-28k uses audio--text evidence. Results are paired with the backbone-only model under the same split and graph inputs.
	
	\begin{table}[t]
		\centering
		{\small
			\setlength{\tabcolsep}{1mm}
			\begin{tabular}{@{}llccc@{}}
				\toprule
				Dataset & Method & Acc. & F1$_+$ & AUC \\
				\midrule
				TwiBot-20 & Backbone & 0.7891 & 0.8078 & 0.8666 \\
				& \fullmethod{} & \textbf{0.8438} & \textbf{0.8659} & \textbf{0.9196} \\
				\addlinespace[1pt]
				Cresci-15 & Backbone & 0.8690 & 0.8883 & 0.9719 \\
				& \fullmethod{} & \textbf{0.9035} & \textbf{0.9195} & \textbf{0.9752} \\
				\addlinespace[1pt]
				YelpZip & Backbone & 0.8182 & 0.5526 & 0.8689 \\
				& \fullmethod{} & \textbf{0.9373} & \textbf{0.7422} & \textbf{0.9354} \\
				\addlinespace[1pt]
				FraudBench & Backbone & 0.8275 & 0.8192 & 0.8438 \\
				& \fullmethod{} & \textbf{0.9780} & \textbf{0.9800} & \textbf{0.9974} \\
				\addlinespace[1pt]
				TeleAntiFraud & Backbone & 0.5806 & 0.7173 & 0.9000 \\
				& \fullmethod{} & \textbf{0.9353} & \textbf{0.9425} & \textbf{0.9936} \\
				\bottomrule
			\end{tabular}
		}
		\caption{Cross-dataset paired results. TwiBot-20, Cresci-2015, and TeleAntiFraud average 13 backbones; YelpZip uses DeepSeek-V4-Pro; FraudBench averages six graphs.}
		\label{tab:cross-dataset-main}
	\end{table}
	
	Accuracy gains range from 3.45 pp on Cresci-2015 to 35.47 pp on TeleAntiFraud; AUC rises by 15.36 pp on FraudBench and 9.36 pp on TeleAntiFraud. These results show that \fullmethod{} can construct evidence chains from different feature sources and use their complementary information to improve fraud detection. Full details are provided in Supplementary Sections S5--S7.
	
	\subsection{Transfer to Unseen Backbones}
	
	Table~\ref{tab:mgtab-unseen-backbone} reuses the discovery router/prior on held-out generic and fraud-specific backbones. None contributes to hard-region discovery, so this experiment separates transferable evidence from co-adaptation to the discovery set.
	
	\begin{table}[t]
		\centering
		{\small
			\setlength{\tabcolsep}{0.5mm}
			\begin{tabular}{@{}lcccccc@{}}
				\toprule
				Backbone & \multicolumn{3}{c}{Backbone only} & \multicolumn{3}{c}{\fullmethod{}} \\
				\cmidrule(lr){2-4}\cmidrule(lr){5-7}
				& Acc. & F1$_{+}$ & AUC & Acc. & F1$_{+}$ & AUC \\
				\midrule
				SGC & 0.8105 & 0.7140 & 0.9109 & \textbf{0.9176} & \textbf{0.8612} & \textbf{0.9783} \\
				GraphTrans. & 0.8928 & 0.8202 & 0.9569 & \textbf{0.9294} & \textbf{0.8764} & \textbf{0.9782} \\
				FiLM & 0.8948 & 0.8193 & 0.9570 & \textbf{0.9301} & \textbf{0.8772} & \textbf{0.9786} \\
				ResGatedGCN & 0.8484 & 0.7588 & 0.9234 & \textbf{0.9275} & \textbf{0.8686} & \textbf{0.9798} \\
				CARE-GNN & 0.9059 & 0.8393 & 0.9637 & \textbf{0.9255} & \textbf{0.8713} & \textbf{0.9779} \\
				PC-GNN & 0.8974 & 0.8242 & 0.9600 & \textbf{0.9301} & \textbf{0.8785} & \textbf{0.9784} \\
				\midrule
				Mean & 0.8749 & 0.7960 & 0.9453 & \textbf{0.9267} & \textbf{0.8722} & \textbf{0.9780} \\
				\bottomrule
			\end{tabular}
		}
		\caption{MGTAB transfer to unseen GNN backbones.}
		\label{tab:mgtab-unseen-backbone}
	\end{table}
	
	\fullmethod{} improves every held-out backbone on all metrics, raising mean Accuracy/F1/AUC by 5.18/7.62/3.27 pp. The set spans graph filtering \cite{sgc}, attention-based propagation \cite{graphtransformer}, feature-wise modulation \cite{film}, gated convolution \cite{resgatedgcn}, CARE-GNN \cite{dou2020enhancing}, and PC-GNN \cite{liu2021pcgnn}. Accuracy also rises by 1.96 pp on CARE-GNN and 3.27 pp on PC-GNN, supporting the enhancer positioning rather than a claim to replace these detectors. Detailed results for the matched routed-fusion controls are provided in Supplementary Section S1.1.
	
	\subsection{GNN-Common Hard Regions}
	\label{sec:hard-region-diagnosis}
	
	Table~\ref{tab:hard-region-summary} uses 13-backbone training OOF predictions except YelpZip, whose diagnosis is post-hoc. Each target backbone is excluded from its induced hard/easy split.
	
	\begin{table}[!ht]
		\centering
		\begin{tabular}{@{}lccc@{}}
			\toprule
			Dataset & Diag. hard rate & Acc. gap & AUC gap \\
			\midrule
			TwiBot-20 & 48.0\% & 0.462 & 0.519 \\
			MGTAB & 18.0\% & 0.424 & 0.445 \\
			Cresci-2015 & 2.7\% & 0.171 & 0.111 \\
			YelpZip & 28.6\% & 0.245 & 0.184 \\
			FraudBench & 37.2\% & 0.559 & 0.781 \\
			TeleAntiFraud & 24.0\% & 0.632 & 0.770 \\
			\bottomrule
		\end{tabular}
		\caption{Cross-dataset GNN common hard-region diagnosis. The diagnostic hard rate is the fraction of nodes marked consensus hard in the leave-one-backbone-out diagnosis.}
		\label{tab:hard-region-summary}
	\end{table}
	
	All six datasets show positive easy--hard gaps, averaging 0.416 in Accuracy and 0.468 in AUC. This supports a shared rather than backbone-specific failure region, while the different hard rates argue against one universal error heuristic: the virtual-node interface is fixed, but routing and evidence models remain replaceable. Supplementary Sections S2, S6, and S7 give task-level diagnostics.
	
	\subsection{Feature Isolation and Final Input Controls}
	
	Table~\ref{tab:mgtab-final-input-ablation-main} compares structural injection with matched score-only inputs. Stacking w/o VN receives the same backbone, prior, router, and mask information but removes virtual-node message passing and the augmented GNN score, isolating structure from post-hoc combination.
	
	\begin{table}[!ht]
		\centering
		{\small
			\setlength{\tabcolsep}{1mm}
			\begin{tabular}{@{}lccc@{}}
				\toprule
				Final input / control & Acc. & F1$_{+}$ & AUC \\
				\midrule
				\fullmethod{} & \textbf{0.9283} & \textbf{0.8710} & \textbf{0.9767} \\
				Stacking w/o VN & 0.9175 & 0.8600 & 0.9755 \\
				w/o augmented score & 0.9166 & 0.8572 & 0.9731 \\
				w/o router score/mask & 0.9164 & 0.8560 & 0.9724 \\
				w/o backbone score & 0.9150 & 0.8552 & 0.9728 \\
				w/o prior score & 0.9122 & 0.8508 & 0.9706 \\
				\bottomrule
			\end{tabular}
		}
		\caption{MGTAB final input ablation and no-VN stacking control.}
		\label{tab:mgtab-final-input-ablation-main}
	\end{table}
	
	All removals reduce performance. \fullmethod{} exceeds matched stacking without virtual nodes by 1.08 pp in Accuracy and 1.10 pp in F1; removing the prior score produces larger drops of 1.61 and 2.02 pp. Thus, learned score combination explains part, but not all, of the gain.
	
	\section{Limitations}
	
	MGTAB has paired tests and a five-seed finalizer check, but full pipeline multi-seed reruns remain expensive. Our evidence processors are replaceable, but current experiments cover selected tabular, language, vision, and audio encoders rather than every model family. FraudBench retains highly separable generated text, while TeleAntiFraud is synthetic and transcript-dominated.
	
	\section{Conclusion}
	
	We presented \fullmethod{}, a model-agnostic enhancer that learns hard regions shared across GNNs and injects feature-isolated, multimodal evidence through source-only virtual class nodes. Across bot, review, refund, and telecom fraud tasks, it consistently strengthens generic, fraud-specific, and unseen GNNs, with particularly clear gains on shared hard nodes and over matched fusion and no-VN stacking controls. These results show that structured, textual, visual, and acoustic evidence can strengthen existing graph detectors without changing their input interface.
   Ultimately, \fullmethod{} serves as a powerful, plug-and-play framework that translates diverse multimodal evidence into targeted structural interventions, directly resolving the blind spots of existing graph detectors.

	\clearpage
\raggedbottom
\setcounter{secnumdepth}{2}
\setcounter{section}{0}
\setcounter{subsection}{0}
\setcounter{table}{0}
\setcounter{figure}{0}
\setcounter{equation}{0}
\renewcommand{\thesection}{S\arabic{section}}
\renewcommand{\thesubsection}{\thesection.\arabic{subsection}}
\renewcommand{\thetable}{S\arabic{table}}
\renewcommand{\thefigure}{S\arabic{figure}}
\renewcommand{\theequation}{S\arabic{equation}}
\section*{Supplementary Material}
This supplementary material is organized as an evidence trail for the main paper. All experiments follow the same strict feature isolation rule as the main paper: GNN backbones receive only data-side node features and graph structure, while evidence chain features are used only by the hard/easy router and the routed prior generator.

\paragraph{Reader's map.}
Section~\ref{sec:supp-detailed-results} reports the MGTAB unseen-backbone controls and the additional TwiBot-20, Cresci-2015, and YelpZip results cited by the main paper. Section~\ref{sec:supp-hard-region} gives the MGTAB and TwiBot-20 hard-region diagnostics. Section~\ref{sec:supp-mgtab-ablation} contains the MGTAB stability and sensitivity analyses. Section~\ref{sec:supp-algorithm} specifies leakage controls, evidence-chain construction, implementation settings, run counts, and the release plan. Section~\ref{sec:supp-yelpzip} records the YelpZip LLM prompt, and Sections~\ref{sec:supp-fraudbench}--\ref{sec:supp-teleantifraud} report the image--text and audio--text experiments.

\section{Detailed Model-Agnostic Results}
\label{sec:supp-detailed-results}

The main paper places the MGTAB 13-backbone result and the MGTAB unseen backbone transfer table in the body because MGTAB is our primary evaluation setting. This section first adds baseline controls for the MGTAB unseen-backbone transfer experiment, and then reports the additional dataset-level result tables that are cited from the main paper.

\subsection{MGTAB Unseen Backbone Baseline Controls}
\label{sec:supp-mgtab-unseen-baseline-controls}

Table~\ref{tab:supp-mgtab-unseen-baseline-controls} extends the main-paper MGTAB \cite{shi2023mgtab} unseen-backbone transfer table by adding routed fusion and score-only stacking controls for the same held-out backbones: SGC \cite{sgc}, Graph Transformer \cite{graphtransformer}, GNN-FiLM \cite{film}, ResGatedGCN \cite{resgatedgcn}, CARE-GNN \cite{dou2020enhancing}, and PC-GNN \cite{liu2021pcgnn}.

\noindent\begin{minipage}{\linewidth}
	\centering
	\small
	\setlength{\tabcolsep}{2.2pt}
	\begin{tabular}{@{}l l ccc@{}}
		\toprule
		Backbone & Method & Acc. & F1+ & AUC \\
		\midrule
		SGC & Backbone only & 0.8105 & 0.7140 & 0.9109 \\
		SGC & Prob. fusion & 0.9020 & 0.8355 & 0.9706 \\
		SGC & Logit fusion & 0.9033 & 0.8363 & 0.9700 \\
		SGC & Stacking w/o VN & 0.9170 & 0.8603 & 0.9712 \\
		SGC & \fullmethod{} & \textbf{0.9176} & \textbf{0.8612} & \textbf{0.9783} \\
		\midrule
		GraphTransformer & Backbone only & 0.8928 & 0.8202 & 0.9569 \\
		GraphTransformer & Prob. fusion & 0.9255 & 0.8710 & 0.9762 \\
		GraphTransformer & Logit fusion & 0.9248 & 0.8698 & 0.9778 \\
		GraphTransformer & Stacking w/o VN & 0.9176 & 0.8597 & 0.9719 \\
		GraphTransformer & \fullmethod{} & \textbf{0.9294} & \textbf{0.8764} & \textbf{0.9782} \\
		\midrule
		FiLM & Backbone only & 0.8948 & 0.8193 & 0.9570 \\
		FiLM & Prob. fusion & 0.9288 & 0.8754 & 0.9775 \\
		FiLM & Logit fusion & 0.9261 & 0.8700 & 0.9783 \\
		FiLM & Stacking w/o VN & 0.9131 & 0.8477 & 0.9659 \\
		FiLM & \fullmethod{} & \textbf{0.9301} & \textbf{0.8772} & \textbf{0.9786} \\
		\midrule
		ResGatedGCN & Backbone only & 0.8484 & 0.7588 & 0.9234 \\
		ResGatedGCN & Prob. fusion & 0.9235 & 0.8675 & 0.9756 \\
		ResGatedGCN & Logit fusion & 0.9229 & 0.8662 & 0.9755 \\
		ResGatedGCN & Stacking w/o VN & 0.9163 & 0.8593 & 0.9773 \\
		ResGatedGCN & \fullmethod{} & \textbf{0.9275} & \textbf{0.8686} & \textbf{0.9798} \\
		\midrule
		CARE-GNN & Backbone only & 0.9059 & 0.8393 & 0.9637 \\
		CARE-GNN & Prob. fusion & 0.9155 & 0.8607 & 0.9684 \\
		CARE-GNN & Logit fusion & 0.9175 & 0.8637 & 0.9694 \\
		CARE-GNN & Stacking w/o VN & 0.9150 & 0.8533 & 0.9710 \\
		CARE-GNN & \fullmethod{} & \textbf{0.9255} & \textbf{0.8713} & \textbf{0.9779} \\
		\midrule
		PC-GNN & Backbone only & 0.8974 & 0.8242 & 0.9600 \\
		PC-GNN & Prob. fusion & 0.9281 & 0.8744 & 0.9779 \\
		PC-GNN & Logit fusion & 0.9281 & 0.8747 & 0.9784 \\
		PC-GNN & Stacking w/o VN & 0.9144 & 0.8533 & 0.9731 \\
		PC-GNN & \fullmethod{} & \textbf{0.9301} & \textbf{0.8785} & \textbf{0.9784} \\
		\midrule
		Mean & Backbone only & 0.8749 & 0.7960 & 0.9453 \\
		Mean & Prob. fusion & 0.9222 & 0.8658 & 0.9760 \\
		Mean & Logit fusion & 0.9221 & 0.8651 & 0.9766 \\
		Mean & Stacking w/o VN & 0.9156 & 0.8556 & 0.9729 \\
		Mean & \fullmethod{} & \textbf{0.9267} & \textbf{0.8722} & \textbf{0.9780} \\
		\bottomrule
	\end{tabular}
	\normalsize
	\captionof{table}{MGTAB unseen-backbone baseline controls. The router and routed prior are trained from the 13 discovery backbones and reused for held-out general GNN and fraud-detector backbones. Fusion and stacking baselines are intentionally coarse, lightweight controls; they do not construct virtual nodes or train an augmented GNN.}
	\label{tab:supp-mgtab-unseen-baseline-controls}
\end{minipage}

The results show that the transferred evidence prior is strong enough to improve simple routed fusion, but \fullmethod{} gives the best mean performance across all three metrics. The comparison against score only stacking without virtual nodes is especially important: it uses the same backbone score, routed prior score, router score, and hard route indicator, yet removing virtual node injection lowers the mean accuracy, F1$_+$, and AUC. This supports the claim that the virtual node augmented GNN contributes beyond post-hoc score stacking.
\subsection{Additional Dataset-Level Results}
\label{sec:supp-additional-dataset-results}

\noindent\begin{minipage}{\linewidth}
\centering
\small
\setlength{\tabcolsep}{1.5pt}
\begin{tabular}{@{}lccc@{}}
\toprule
Method & Accuracy & F1$_{+}$ & AUC \\
\midrule
Backbone only & 0.7891 & 0.8078 & 0.8666 \\
GNN probability ensemble & 0.8233 & 0.8439 & 0.9063 \\
Weighted probability fusion & 0.8399 & 0.8577 & 0.9174 \\
Weighted logit fusion & 0.8399 & 0.8577 & 0.9176 \\
\fullmethod{} & \textbf{0.8438} & \textbf{0.8659} & \textbf{0.9196} \\
\bottomrule
\end{tabular}
\normalsize
\captionof{table}{TwiBot-20 test performance averaged over 13 GNN backbones.}
\label{tab:supp-twibot20-main}
\end{minipage}

\noindent\begin{minipage}{\linewidth}
\centering
\small
\setlength{\tabcolsep}{3.0pt}
\begin{tabular}{@{}lccc@{}}
\toprule
Method & Accuracy & F1$_{+}$ & AUC \\
\midrule
Backbone only & 0.8690 & 0.8883 & 0.9719 \\
GNN probability ensemble & 0.8737 & 0.8928 & 0.9599 \\
Weighted probability fusion & 0.8704 & 0.8896 & 0.9718 \\
Weighted logit fusion & 0.8690 & 0.8883 & 0.9711 \\
\fullmethod{} & \textbf{0.9035} & \textbf{0.9195} & \textbf{0.9752} \\
\bottomrule
\end{tabular}
\normalsize
\captionof{table}{Cresci-2015 test performance averaged over 13 GNN backbones.}
\label{tab:supp-cresci2015-main}
\end{minipage}

The YelpZip \cite{rayana2015collective} experiment is not intended to show that an LLM prior alone is a strong standalone detector. Instead, it tests whether natural-language evidence chains can instantiate the routed prior generator in \fullmethod{}. The routed LLM is either Qwen-Plus \cite{alibaba2026qwenplus} or DeepSeek-V4-Pro \cite{deepseek2026v4pro}. The prior only rows are diagnostic: easy route reviews receive the neutral score $0.5$, and the raw LLM scores on routed hard reviews are noisy. Even under this weak prior setting, \fullmethod{} extracts useful aligned evidence and improves the final prediction over backbone only and routed fusion. Compared with score-only stacking without virtual nodes, the main improvement is in AUC, while accuracy changes only slightly; this suggests that virtual-node injection improves hard review ranking/calibration rather than merely shifting the operating point.

\noindent\begin{minipage}{\linewidth}
\centering
\small
\setlength{\tabcolsep}{3.0pt}
\begin{tabular}{@{}llccc@{}}
\toprule
Prior & Method & Acc. & F1$_+$ & AUC \\
\midrule
-- & Backbone only & 0.8182 & 0.5526 & 0.8689 \\
\midrule
DeepSeek & Prior only & 0.3096 & 0.2368 & 0.5246 \\
DeepSeek & Weighted prob. fusion & 0.8892 & 0.7190 & 0.8948 \\
DeepSeek & Weighted logit fusion & 0.8889 & 0.7191 & 0.8920 \\
DeepSeek & Stacking w/o VN & 0.9357 & 0.7223 & 0.8937 \\
DeepSeek & Evi-VN & \textbf{0.9373} & \textbf{0.7422} & \textbf{0.9354} \\
\midrule
Qwen & Prior only & 0.2543 & 0.2294 & 0.4883 \\
Qwen & Weighted prob. fusion & 0.8473 & 0.6311 & 0.8902 \\
Qwen & Weighted logit fusion & 0.8458 & 0.6259 & 0.8884 \\
Qwen & Stacking w/o VN & 0.9341 & 0.7180 & 0.8964 \\
Qwen & Evi-VN & \textbf{0.9354} & \textbf{0.7349} & \textbf{0.9288} \\
\bottomrule
\end{tabular}
\normalsize
\captionof{table}{YelpZip LLM prior comparison.}
\label{tab:yelpzip-llm20-method-summary}
\end{minipage}

The shuffled control separates evidence alignment from the marginal score distribution. Real DeepSeek priors improve \fullmethod{} over shuffled priors by 0.0060 accuracy, 0.0203 F1$_+$, and 0.0135 AUC; real Qwen priors improve it by 0.0054 accuracy, 0.0239 F1$_+$, and 0.0159 AUC. These drops under shuffled priors show that the true LLM prior is not ignored by the pipeline. Thus the YelpZip result should be read as evidence that \fullmethod{} can use weak natural language priors through hard routing and source-only virtual-node injection, not as evidence that the LLM prior alone solves YelpZip.

\noindent\begin{minipage}{\linewidth}
\centering
\small
\setlength{\tabcolsep}{3.0pt}
\begin{tabular}{@{}llccc@{}}
\toprule
Prior source & Method & Acc. & F1$_+$ & AUC \\
\midrule
Real DeepSeek & Evi-VN & \textbf{0.9373} & \textbf{0.7422} & \textbf{0.9354} \\
Shuffled DeepSeek & Evi-VN & 0.9313 & 0.7219 & 0.9219 \\
\midrule
Real Qwen & Evi-VN & \textbf{0.9354} & \textbf{0.7349} & \textbf{0.9288} \\
Shuffled Qwen & Evi-VN & 0.9300 & 0.7110 & 0.9129 \\
\bottomrule
\end{tabular}
\normalsize
\captionof{table}{YelpZip full-pipeline shuffled-prior control. The shuffled control keeps the same routed hard nodes and the same LLM score distribution, but permutes scores before virtual-edge construction and score finalization.}
\label{tab:yelpzip-shuffled-prior-control}
\end{minipage}

\noindent\begin{minipage}{\linewidth}
\centering
\small
\setlength{\tabcolsep}{2.5pt}
\begin{tabular}{@{}lcccccc@{}}
\toprule
Backbone & \multicolumn{3}{c}{Backbone only} & \multicolumn{3}{c}{Evi-VN} \\
\cmidrule(lr){2-4}\cmidrule(l){5-7}
& Acc. & F1$_+$ & AUC & Acc. & F1$_+$ & AUC \\
\midrule
MLP & 0.7278 & 0.3966 & 0.7928 & 0.8989 & 0.6001 & 0.8390 \\
GCN & 0.8934 & 0.5997 & 0.9078 & 0.9469 & 0.7809 & 0.9587 \\
GAT & 0.8784 & 0.5869 & 0.9011 & 0.9449 & 0.7710 & 0.9629 \\
GraphSAGE & 0.8973 & 0.6256 & 0.9188 & 0.9415 & 0.7527 & 0.9302 \\
GCNII & 0.8776 & 0.5720 & 0.8936 & 0.9374 & 0.7449 & 0.9336 \\
APPNP & 0.1297 & 0.2297 & 0.5106 & 0.8921 & 0.5436 & 0.8311 \\
GPRGNN & 0.8641 & 0.5099 & 0.8678 & 0.9319 & 0.7186 & 0.9498 \\
MixHop & 0.8849 & 0.5807 & 0.9018 & 0.9465 & 0.7829 & 0.9629 \\
SIGN & 0.9011 & 0.5976 & 0.9027 & 0.9383 & 0.7435 & 0.9426 \\
ChebNet & 0.9038 & 0.6346 & 0.9297 & 0.9523 & 0.8032 & 0.9456 \\
GraphConv & 0.8726 & 0.5505 & 0.8941 & 0.9507 & 0.7968 & 0.9731 \\
TAGCN & 0.8992 & 0.6466 & 0.9330 & 0.9543 & 0.8121 & 0.9715 \\
GIN & 0.9061 & 0.6529 & 0.9414 & 0.9494 & 0.7977 & 0.9599 \\
\midrule
Mean & 0.8182 & 0.5526 & 0.8689 & \textbf{0.9373} & \textbf{0.7422} & \textbf{0.9354} \\
\bottomrule
\end{tabular}
\normalsize
\captionof{table}{YelpZip per-backbone results using DeepSeek-V4-Pro as the routed prior generator.}
\label{tab:yelpzip-deepseek-backbone}
\end{minipage}

\noindent\begin{minipage}{\linewidth}
\centering
\small
\setlength{\tabcolsep}{2.5pt}
\begin{tabular}{@{}lcccccc@{}}
\toprule
Backbone & \multicolumn{3}{c}{Backbone only} & \multicolumn{3}{c}{Evi-VN} \\
\cmidrule(lr){2-4}\cmidrule(l){5-7}
& Acc. & F1$_+$ & AUC & Acc. & F1$_+$ & AUC \\
\midrule
MLP & 0.7278 & 0.3966 & 0.7928 & 0.8980 & 0.6001 & 0.8423 \\
GCN & 0.8934 & 0.5997 & 0.9078 & 0.9417 & 0.7580 & 0.9484 \\
GAT & 0.8784 & 0.5869 & 0.9011 & 0.9498 & 0.7833 & 0.9492 \\
GraphSAGE & 0.8973 & 0.6256 & 0.9188 & 0.9432 & 0.7608 & 0.9322 \\
GCNII & 0.8776 & 0.5720 & 0.8936 & 0.9390 & 0.7381 & 0.9238 \\
APPNP & 0.1297 & 0.2297 & 0.5106 & 0.8944 & 0.5619 & 0.8174 \\
GPRGNN & 0.8641 & 0.5099 & 0.8678 & 0.9311 & 0.7145 & 0.9391 \\
MixHop & 0.8849 & 0.5807 & 0.9018 & 0.9403 & 0.7576 & 0.9537 \\
SIGN & 0.9011 & 0.5976 & 0.9027 & 0.9356 & 0.7329 & 0.9346 \\
ChebNet & 0.9038 & 0.6346 & 0.9297 & 0.9524 & 0.7999 & 0.9472 \\
GraphConv & 0.8726 & 0.5505 & 0.8941 & 0.9336 & 0.7416 & 0.9587 \\
TAGCN & 0.8992 & 0.6466 & 0.9330 & 0.9519 & 0.8029 & 0.9675 \\
GIN & 0.9061 & 0.6529 & 0.9414 & 0.9493 & 0.8015 & 0.9602 \\
\midrule
Mean & 0.8182 & 0.5526 & 0.8689 & \textbf{0.9354} & \textbf{0.7349} & \textbf{0.9288} \\
\bottomrule
\end{tabular}
\normalsize
\captionof{table}{YelpZip per-backbone results using Qwen-Plus as the routed prior generator.}
\label{tab:yelpzip-qwen-backbone}
\end{minipage}

\section{Consensus Hard Region Diagnostics}
\label{sec:supp-hard-region}

This section expands the cross dataset hard-region summary in the main paper. For MGTAB and TwiBot-20, hard/easy labels are computed from training node OOF predictions of the discovery backbones. A node receives a hard vote from a backbone if the OOF prediction is wrong or close to the decision boundary, $|p_i^{G,\OOF}-0.5|\le 0.20$. Leave-one-backbone-out diagnostics exclude the evaluated target backbone from hard region discovery, so the resulting easy--hard gap measures transfer of the blind-spot definition rather than memorization of one model's errors.

\subsection{MGTAB Diagnostics}

\noindent\begin{minipage}{\linewidth}
\centering
\small
\setlength{\tabcolsep}{2.5pt}
\begin{tabular}{@{}lcccccc@{}}
\toprule
Backbone & H-Acc. & E-Acc. & Gap & H-F1 & E-F1 & AUC gap \\
\midrule
MLP & 0.547 & 0.978 & 0.431 & 0.459 & 0.957 & 0.413 \\
GCN & 0.407 & 0.916 & 0.509 & 0.355 & 0.850 & 0.579 \\
GAT & 0.422 & 0.860 & 0.439 & 0.368 & 0.766 & 0.542 \\
GraphSAGE & 0.549 & 0.992 & 0.442 & 0.447 & 0.983 & 0.428 \\
GCNII & 0.518 & 0.992 & 0.474 & 0.468 & 0.985 & 0.449 \\
APPNP & 0.450 & 0.622 & 0.172 & 0.432 & 0.497 & 0.226 \\
GPRGNN & 0.533 & 0.986 & 0.453 & 0.453 & 0.972 & 0.445 \\
MixHop & 0.437 & 0.915 & 0.479 & 0.374 & 0.846 & 0.550 \\
SIGN & 0.560 & 0.990 & 0.429 & 0.459 & 0.979 & 0.424 \\
ChebNet & 0.559 & 0.983 & 0.424 & 0.461 & 0.967 & 0.414 \\
GraphConv & 0.447 & 0.859 & 0.412 & 0.343 & 0.748 & 0.433 \\
TAGCN & 0.526 & 0.993 & 0.467 & 0.474 & 0.986 & 0.431 \\
GIN & 0.461 & 0.841 & 0.380 & 0.321 & 0.702 & 0.456 \\
\bottomrule
\end{tabular}
\normalsize
\captionof{table}{MGTAB leave-one-backbone-out blind-spot test. Hard nodes are defined using the other 12 backbones, and the held out target backbone is evaluated on the resulting hard/easy split.}
\label{tab:mgtab-loo-hard-details}
\end{minipage}

\noindent\begin{minipage}{\linewidth}
\centering
\small
\setlength{\tabcolsep}{3.0pt}
\begin{tabular}{@{}lccc@{}}
\toprule
Detector & Accuracy & F1 & AUC \\
\midrule
Logistic Regression & 0.733 & 0.492 & 0.800 \\
Random Forest & 0.801 & 0.519 & 0.849 \\
ExtraTrees & 0.739 & 0.512 & 0.835 \\
\bottomrule
\end{tabular}
\normalsize
\captionof{table}{MGTAB hard/easy separability using simple detectors trained on node features and graph statistics. The positive class is consensus hard.}
\label{tab:mgtab-hard-easy-separability}
\end{minipage}

The MGTAB appendix supports the main diagnosis: every target backbone performs worse on the leave-one-backbone-out hard region, with mean accuracy and AUC gaps of 0.424 and 0.445. This region is also structurally meaningful, as a Random Forest predicts consensus-hard membership with 0.849 AUC.

\begin{figure}[t]
\centering
\includegraphics[width=0.95\linewidth]{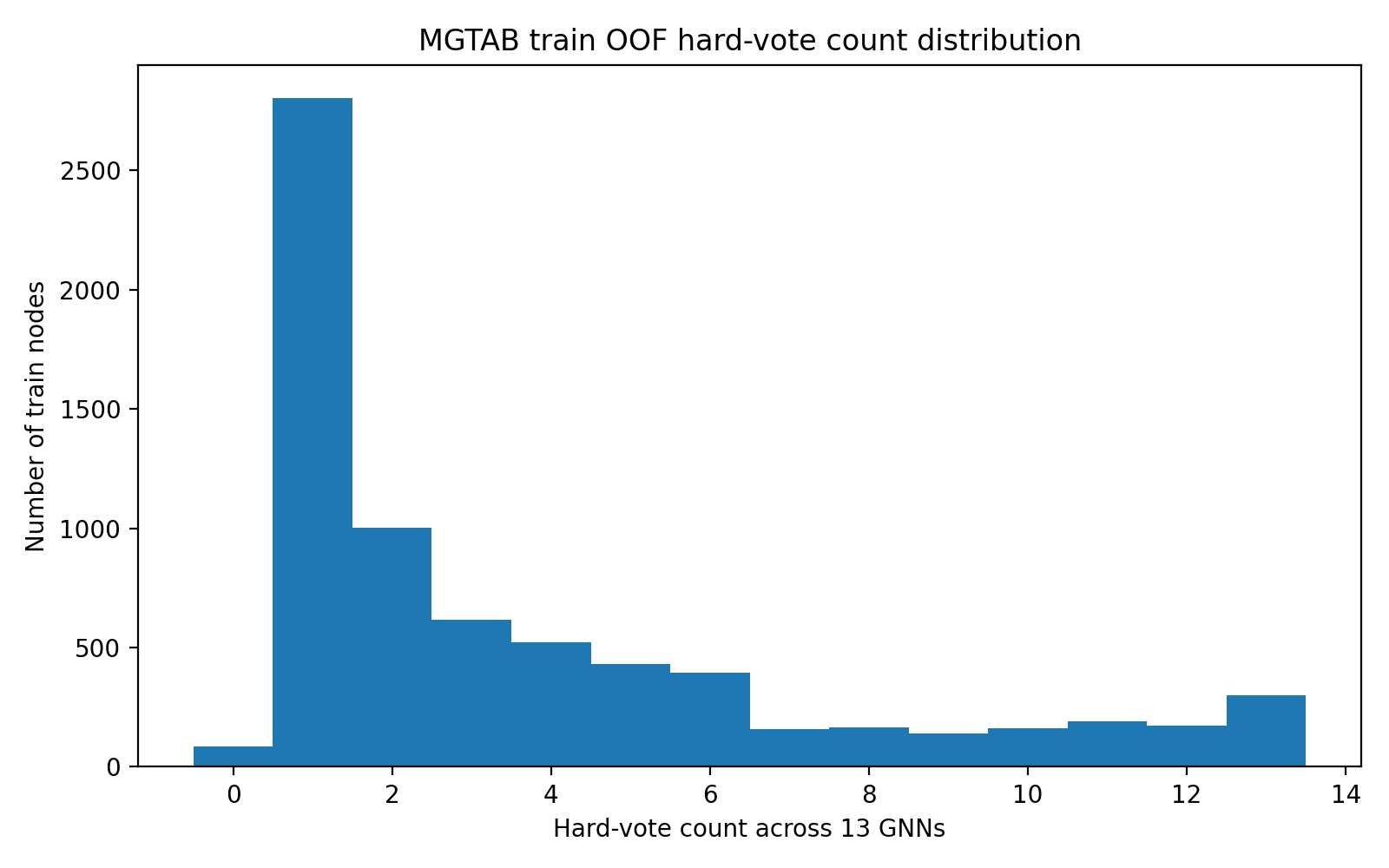}
\caption{MGTAB distribution of hard vote counts over the 13 GNN backbones.}
\label{fig:mgtab-hard-vote-count}
\end{figure}

\begin{figure}[t]
\centering
\includegraphics[width=0.95\linewidth]{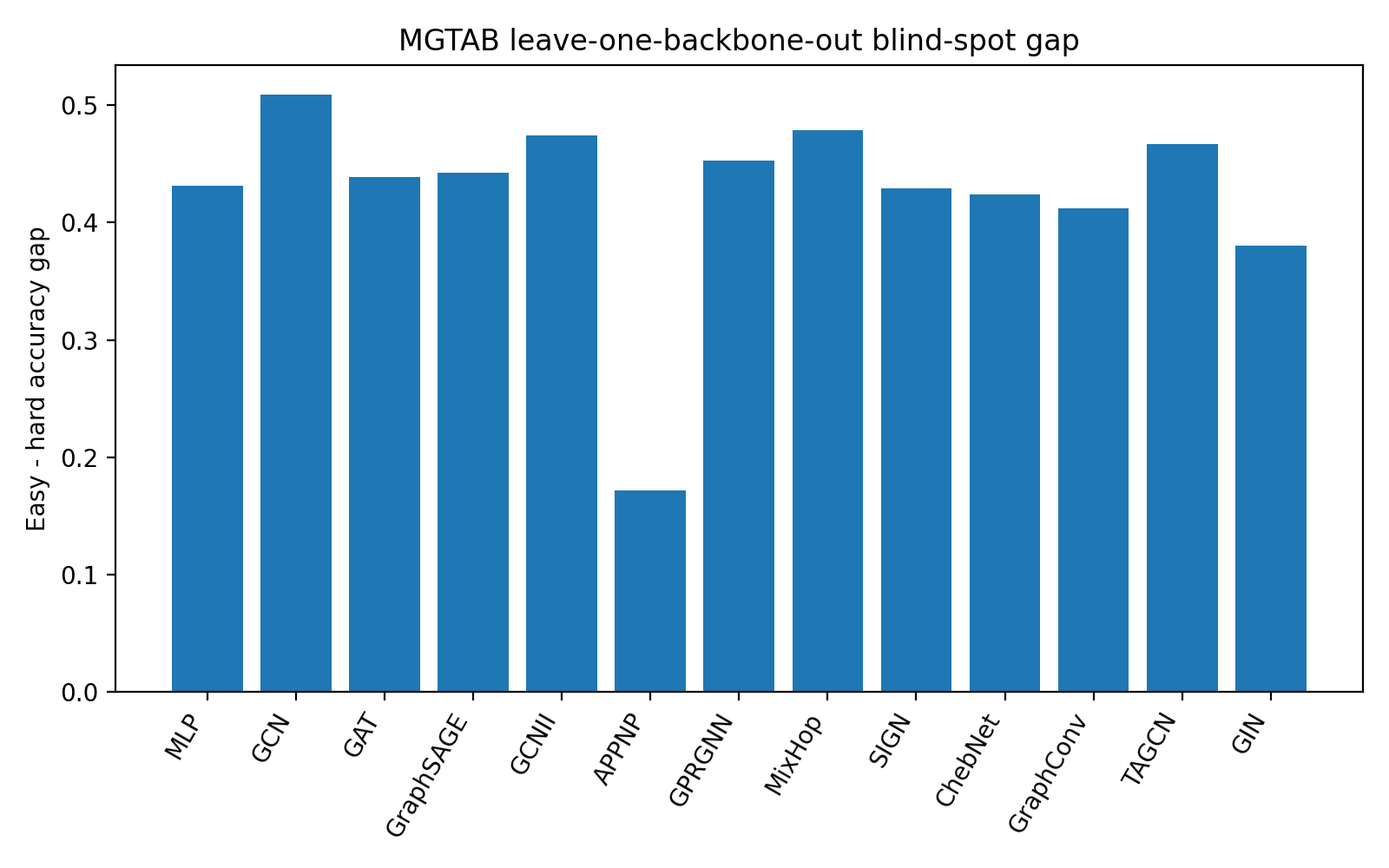}
\caption{MGTAB leave-one-backbone-out easy-hard accuracy gap. The target backbone is excluded from hard region discovery.}
\label{fig:mgtab-loo-acc-gap}
\end{figure}
\subsection{TwiBot-20 Diagnostics}

The TwiBot-20 diagnostics mirror the MGTAB protocol: each held out backbone is excluded from discovery before the induced hard/easy split is evaluated on that target model.

\noindent\begin{minipage}{\linewidth}
\centering
\small
\setlength{\tabcolsep}{2.5pt}
\begin{tabular}{@{}lcccccc@{}}
\toprule
Unseen & H-Acc. & E-Acc. & Gap & H-AUC & E-AUC & Acc. \\
\midrule
SGC & 0.427 & 0.927 & 0.500 & 0.385 & 0.975 & 0.442 \\
GATv2 & 0.438 & 0.929 & 0.491 & 0.395 & 0.978 & 0.451 \\
\bottomrule
\end{tabular}
\normalsize
\captionof{table}{TwiBot-20 unseen diagnostic backbones.}
\label{tab:unseen-hard-transfer}
\end{minipage}

\noindent\begin{minipage}{\linewidth}
\centering
\small
\setlength{\tabcolsep}{2.5pt}
\begin{tabular}{@{}lccccc@{}}
\toprule
Backbone & H-Acc. & E-Acc. & Gap & H-AUC & E-AUC \\
\midrule
MLP & 0.509 & 0.943 & 0.434 & 0.515 & 0.988 \\
GCN & 0.452 & 0.950 & 0.498 & 0.415 & 0.989 \\
GAT & 0.458 & 0.927 & 0.468 & 0.416 & 0.979 \\
GraphSAGE & 0.489 & 0.988 & 0.499 & 0.465 & 0.998 \\
GCNII & 0.472 & 0.972 & 0.499 & 0.449 & 0.996 \\
APPNP & 0.458 & 0.686 & 0.228 & 0.459 & 0.819 \\
GPRGNN & 0.503 & 0.986 & 0.483 & 0.483 & 0.998 \\
MixHop & 0.453 & 0.958 & 0.505 & 0.426 & 0.991 \\
SIGN & 0.477 & 0.989 & 0.511 & 0.452 & 0.999 \\
ChebNet & 0.506 & 0.970 & 0.465 & 0.479 & 0.995 \\
GraphConv & 0.487 & 0.987 & 0.500 & 0.461 & 0.998 \\
TAGCN & 0.495 & 0.993 & 0.498 & 0.473 & 0.998 \\
GIN & 0.466 & 0.884 & 0.418 & 0.458 & 0.951 \\
\bottomrule
\end{tabular}
\normalsize
\captionof{table}{TwiBot-20 leave-one-backbone-out blind spot test. Hard nodes are defined using the other 12 backbones, and the held out target backbone is evaluated on the resulting hard/easy split.}
\label{tab:loo-hard-details}
\end{minipage}

\begin{figure}[t]
\centering
\includegraphics[width=0.95\linewidth]{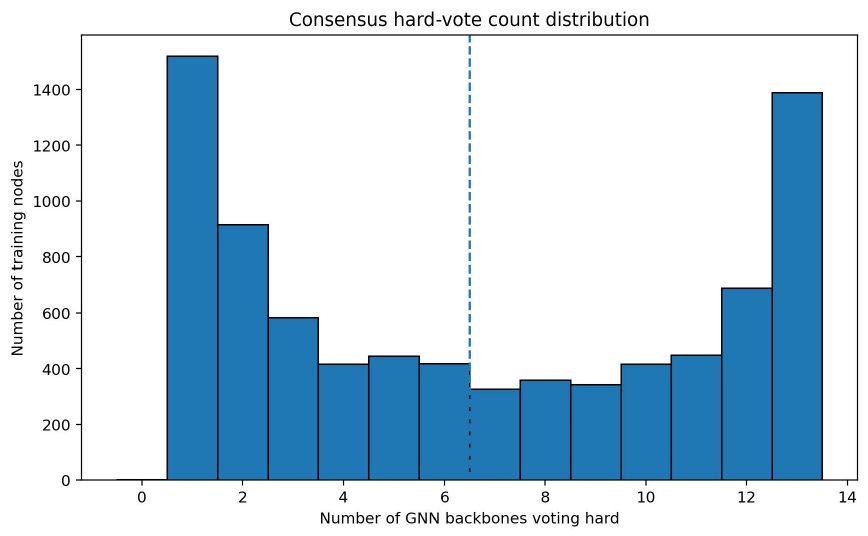}
\caption{TwiBot-20 distribution of hard vote counts over the 13 GNN backbones.}
\label{fig:hard-vote-count}
\end{figure}

\begin{figure}[t]
\centering
\includegraphics[width=0.95\linewidth]{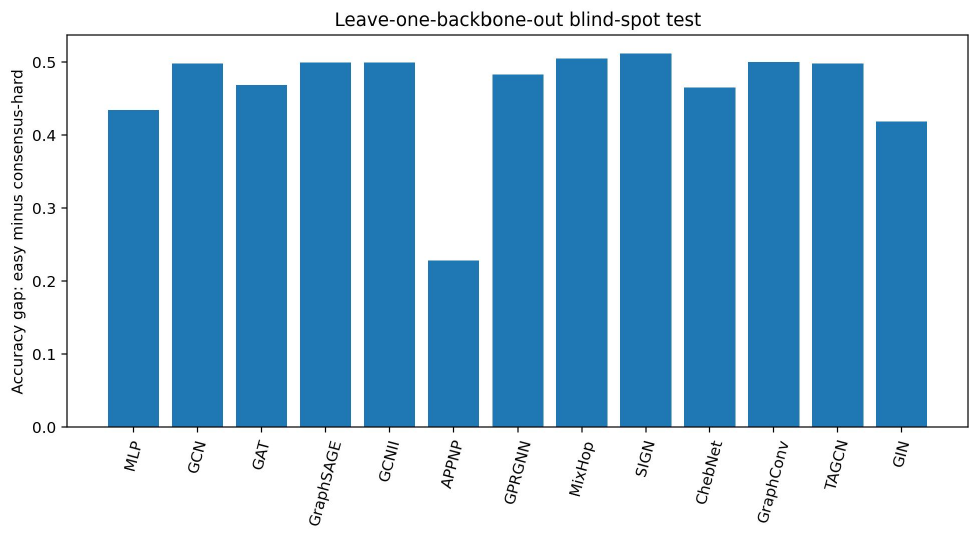}
\caption{TwiBot-20 leave-one-backbone-out easy-hard accuracy gap. The target backbone is excluded from hard region discovery.}
\label{fig:loo-acc-gap}
\end{figure}
\noindent\begin{minipage}{\linewidth}
\centering
\small
\setlength{\tabcolsep}{3.0pt}
\begin{tabular}{@{}lccc@{}}
\toprule
Detector & Accuracy & F1 & AUC \\
\midrule
Logistic & 0.689 & 0.684 & 0.768 \\
Random Forest & 0.686 & 0.650 & 0.772 \\
ExtraTrees & 0.658 & 0.548 & 0.754 \\
\bottomrule
\end{tabular}
\normalsize
\captionof{table}{TwiBot-20 hard/easy separability from node and graph statistics.}
\label{tab:hard-easy-separability}
\end{minipage}

\begin{figure}[t]
\centering
\includegraphics[width=0.95\linewidth]{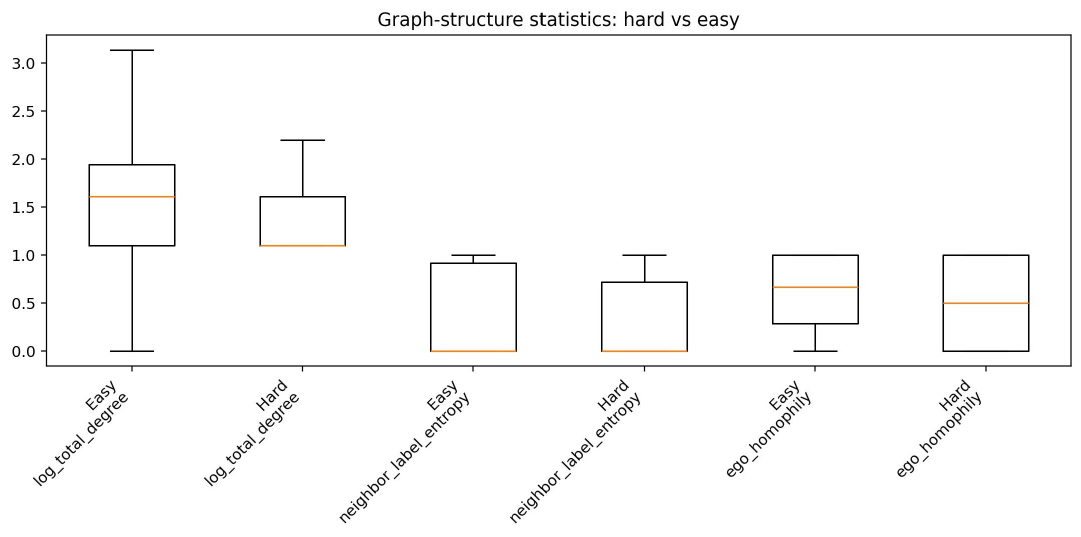}
\caption{TwiBot-20 structural differences between consensus hard and easy nodes.}
\label{fig:hard-easy-graph-stats}
\end{figure}
\section{MGTAB Stability and Sensitivity}
\label{sec:supp-mgtab-ablation}

This section varies the finalizer seed, router threshold, and discovery-backbone count under the current \fullmethod{} protocol. The matched no-VN stacking control and final-input ablation remain in the main paper because they directly test structural injection.

\subsection{Finalizer Seed Stability}

Table~\ref{tab:mgtab-finalizer-seed-stability} varies only the random seed of the score only finalizer while keeping the selected GNN, virtual node, router, and prior caches fixed.

\noindent\begin{minipage}{\linewidth}
\centering
\small
\setlength{\tabcolsep}{2.8pt}
\begin{tabular}{@{}lccc@{}}
\toprule
Finalizer seed & Acc. & F1$_+$ & AUC \\
\midrule
1 & 0.9265 & 0.8682 & 0.9757 \\
7 & 0.9268 & 0.8688 & 0.9756 \\
13 & 0.9271 & 0.8696 & 0.9760 \\
21 & 0.9253 & 0.8661 & 0.9744 \\
42 & 0.9285 & 0.8711 & 0.9765 \\
\midrule
Mean $\pm$ std & 0.9269$\pm$0.0011 & 0.8687$\pm$0.0019 & 0.9756$\pm$0.0008 \\
\bottomrule
\end{tabular}
\normalsize
\captionof{table}{MGTAB score only finalizer seed stability. GNN, virtual node, router, and prior caches are fixed.}
\label{tab:mgtab-finalizer-seed-stability}
\end{minipage}

\subsection{Router Threshold Selection}
\label{sec:supp-router-threshold}

Table~\ref{tab:mgtab-router-threshold} selects $\tau_h$ by final development Accuracy averaged over the 13 MGTAB backbones. The selected threshold is fixed before test evaluation.

\noindent\begin{minipage}{\linewidth}
	\centering
	\small
	\setlength{\tabcolsep}{3.0pt}
	\begin{tabular}{@{}lccccc@{}}
		\toprule
		\multicolumn{6}{c}{Threshold selection} \\
		\midrule
		Router $\tau_h$ & 0.01 & 0.02 & 0.03 & 0.04 & 0.05 \\
		Dev Acc. (\%) & 93.11 & 93.32 & 93.37 & 93.33 & 93.31 \\
		\midrule
		Router $\tau_h$ & 0.06 & \textbf{0.07} & 0.08 & 0.09 & 0.10 \\
		Dev Acc. (\%) & 93.44 & \textbf{93.47} & 93.31 & 93.38 & 93.31 \\
		\bottomrule
	\end{tabular}

	\normalsize
	\captionof{table}{MGTAB router-threshold selection.}
	\label{tab:mgtab-router-threshold}
\end{minipage}

\subsection{Discovery Backbone Count Sensitivity}
\label{sec:supp-discovery-count}

Table~\ref{tab:mgtab-discovery-count} tests whether the hard-region discovery step requires all 13 backbones. We rerun the routed XGBoost prior and source-only virtual-node pipeline with 3, 5, 8, or 13 discovery backbones, while evaluating the same representative GCN/GAT/GIN target backbones. All runs reuse leakage-free OOF discovery caches, retrain the router/prior, and retrain the augmented GNN under the same compact VN grid.

\noindent\begin{minipage}{\linewidth}
\centering
\small
\setlength{\tabcolsep}{3.0pt}
\begin{tabular}{@{}crrc@{}}
\toprule
Disc. B & Cons. hard & Test route & Backbone Acc./F1/AUC \\
\midrule
3  & 32.2\% & 71.3\% & .788/.675/.862 \\
5  & 18.4\% & 70.6\% & .788/.675/.862 \\
8  & 30.7\% & 71.2\% & .788/.675/.862 \\
13 & 18.0\% & 70.8\% & .788/.675/.862 \\
\bottomrule
\end{tabular}

\begin{tabular}{@{}ccc@{}}
\toprule
Disc. B & \fullmethod{} Acc./F1/AUC & $\Delta$ Acc./F1/AUC \\
\midrule
3  & .911/.840/.963 & +.123/+.165/+.101 \\
5  & .897/.821/.949 & +.109/+.146/+.087 \\
8  & .912/.846/.963 & +.124/+.172/+.101 \\
13 & .901/.827/.950 & +.113/+.152/+.088 \\
\bottomrule
\end{tabular}
\normalsize
\captionof{table}{MGTAB discovery-backbone-count sensitivity on representative GCN/GAT/GIN target backbones. ``Cons. hard'' is the training consensus-hard rate induced by the discovery set; ``Test route'' is the learned hard-route rate on the test split.}
\label{tab:mgtab-discovery-count}
\end{minipage}

The gains remain positive for every discovery-set size, so 13 backbones are not a hard requirement. We use 13 in the main experiment because it is the largest and most diverse discovery set available in our completed OOF cache, giving the most conservative shared-hard-region definition for the full-backbone main run.
\section{Algorithm and Implementation Details}
\label{sec:supp-algorithm}

\subsection{Leakage Free Training Checklist}

\paragraph{Input.}
The framework receives graph $G=(\V,\E)$, data side GNN node features $\mathbf{X}^{G}$, evidence chain features $\mathbf{S}^{E}$, train/dev/test splits, real node labels on $\T$, graph learner family $\mathcal{F}$, router family $\mathcal{G}$, and prior generator family $\mathcal{H}$.

\paragraph{Router and prior generation.}
Consensus hard labels are constructed from leakage free multi backbone GNN OOF predictions on the training split. The hard/easy router $g_\psi$ estimates these labels from $\mathbf{s}_i^{E}$. Training node router scores are produced out of fold and calibrated on training data only; development and test scores are produced by the full training router and the same calibration map. The route mask $m_i$ determines whether node $i$ enters the prior generator. Easy route nodes bypass $h_\phi$ and receive $\hat q_i=0.5$. Conditional on this route mask, $h_\phi$ is trained on routed hard training samples to predict the task label $y_i$, not the hard/easy label.

\paragraph{Virtual nodes and edges.}
Imputation and standardization are fitted only on training real node data-side features. Two fixed virtual node features are constructed from training set class prototypes with appended identity dimensions. Evidence-chain features are never appended to the GNN node feature matrix. A routed hard node receives a source-only edge from the positive virtual class node if $\hat q_i$ is above the positive threshold and from the negative virtual class node if $\hat q_i$ is below the negative threshold. Priors near $0.5$ produce weak or no structural injection.

\paragraph{Augmented GNN and finalizer.}
The GNN is trained on the augmented graph with supervised losses computed only on real training nodes. \proxyreg{} uses the two virtual nodes as representation level class proxies during GNN training. The finalizer consumes only scalar scores: the backbone probability, augmented GNN probability, calibrated prior, router score, and route mask. It never consumes raw evidence chain features or augmented GNN hidden vectors.

\subsection{Dataset-Specific Evidence Chains}
\label{sec:supp-evidence-chain-definitions}

All six datasets are publicly accessible and downloadable from their cited releases; we therefore do not upload or redistribute raw dataset copies. The original data remain under their source licenses. Below, we report dataset statistics and splits together with the retained evidence features and the label, split, model-output, and shortcut fields excluded to prevent leakage. In this paper, an evidence chain is a dataset-specific non-graph evidence record $\mathbf{s}_i^E$ consumed only by the hard/easy router and routed prior generator. It need not be an LLM rationale: in tabular account datasets it is a structured statistical vector, while in YelpZip it is a compact natural-language evidence record read by an LLM. The GNN input $\mathbf{x}_i^G$ is kept separate from $\mathbf{s}_i^E$ throughout training and inference.

\paragraph{MGTAB.}
MGTAB contains 10,199 labeled Twitter accounts, 788-dimensional account features, and seven relation types. We use a stratified 7,139/1,530/1,530 train/dev/test split. The GNN input is the imputed and standardized released account feature tensor, with relation types collapsed to a homogeneous graph. The evidence chain is account-level tabular/statistical evidence from the same released source and is consumed by the XGBoost router/prior. This is a same-source but feature-isolated setting: the evidence branch may exploit tabular signals, but raw evidence is not appended to the GNN feature matrix or the finalizer. Labels, splits, GNN votes, and hard-region annotations are excluded from $\mathbf{s}_i^E$.

\paragraph{TwiBot-20.}
TwiBot-20 is a Twitter bot benchmark with user features and social relations, evaluated with its official split. The GNN uses the prepared data-side account features and graph relations. The evidence chain contains structured bot-evidence descriptors summarizing profile/account metadata, behavior/activity statistics, and local social-context cues for the router and XGBoost prior. Bot labels, split indicators, target GNN outputs, and discovery hard labels are not included as evidence features.

\paragraph{Cresci-2015.}
Cresci-2015 contains human and bot/fake-follower accounts. Our graph has 5,301 nodes, 39 GNN features, and a 2,990/997/1,314 train/dev/test split. The GNN uses the prepared account-level feature vector and follower/following graph. The evidence chain contains structured account and behavior statistics, including profile, activity, timing, ratio, and local graph-summary cues for the routed prior. Bot-campaign labels and train/dev/test membership are not used as features; hard labels supervise only the router on training OOF diagnostics.

\paragraph{YelpZip.}
YelpZip contains 608,458 reviews and uses a 60/20/20 train/dev/test split. The GNN uses 25 non-LLM review/user/product statistics and the review graph. The evidence chain for routed hard reviews contains review text, star rating, product-month context, reviewer history, burst behavior, and graph-neighbor summaries. Qwen-Plus/DeepSeek-V4-Pro output calibrated fake-review priors and short evidence codes. Prompts explicitly exclude labels, splits, GNN votes, and hard-region annotations. YelpZip is evaluated as a transductive review-graph task: context may use observed graph evidence, but never gold labels.

\paragraph{FraudBench.}
FraudBench is not released with a canonical graph. We construct claim nodes and shared user/item/location edges; the GNN receives only the three corresponding degree features. The evidence chain is a 238-dimensional image--text vector containing a frozen visual embedding, decoded-pixel statistics, and review text features. Review/product ratings, platform/category identity, file-level metadata, labels, splits, and GNN diagnostics are excluded from this vector.

\paragraph{TeleAntiFraud-28k.}
We construct one call node and its utterance nodes from each dialogue, with call--utterance and consecutive-utterance edges. The GNN receives 19 structural and timing features. The evidence chain contains 96-dimensional frozen XLSR-53 audio \cite{conneau2021xlsr} and 128-dimensional multilingual MiniLM transcript representations \cite{wang2020minilm,reimers2019sentencebert}; source identities, labels, splits, and GNN diagnostics are excluded.

\subsection{Experimental Reproducibility}
\label{sec:supp-reproducibility}

\paragraph{Splits and model selection.}
All imputers, scalers, calibrators, thresholds, and learned evidence transforms are fitted on training data only. Hyperparameters and operating thresholds are selected on development data; test labels are used only for final reporting. The primary MGTAB pipeline uses five stratified OOF folds for discovery, routing, and routed-prior training. Training-node inputs to the finalizer are cross-fitted, while development and test predictions are produced by models refitted on the complete training split.

\paragraph{Primary configuration.}
Table~\ref{tab:supp-mgtab-repro-config} lists the locked MGTAB settings used for the main comparison. The primary router and prior generator use XGBoost \cite{chen2016xgboost}, and OOF probabilities use Platt calibration \cite{platt1999probabilistic}. Backbone-specific propagation operators retain their standard definitions; the shared optimization settings below keep the comparison controlled.

\noindent\begin{minipage}{\linewidth}
\centering
\small
\setlength{\tabcolsep}{2.6pt}
\begin{tabular}{@{}p{0.25\linewidth}p{0.68\linewidth}@{}}
\toprule
Component & Locked setting \\
\midrule
Split and seed & Stratified 70/15/15; seed 42 \\
OOF protocol & 5 folds; stratified; train-only preprocessing \\
GNN optimization & Hidden 128; dropout 0.35; Adam; lr $10^{-3}$; wd $10^{-4}$ \\
Early stopping & 350 epochs, patience 45; OOF 220, patience 30 \\
Hard target & Error or $|p_i^{G,\OOF}-0.5|\leq0.20$; 50\% consensus \\
Router/prior & XGBoost, 500 trees each; OOF Platt calibration \\
Routing & Dev-selected $\tau_h$ on $\{0.01,0.02,\ldots,0.10\}$; easy nodes use $\hat q_i=0.5$ \\
Virtual injection & Two class nodes; source-only confidence-weighted edges \\
Proxy regularization & $(\lambda_p,\lambda_s,\lambda_a)=(0.10,0.01,0.003)$ \\
Proxy geometry & $\tau_p=0.20$, $\gamma_{\mathrm{sep}}=0$, $\rho_{\mathrm{anc}}=0.75$, warm-up 30 epochs \\
Final prediction & Score-only finalizer; no residual prior-logit bias \\
Selection and reporting & Development selection; fixed 0.5 test threshold \\
\bottomrule
\end{tabular}
\normalsize
\captionof{table}{Locked configuration for the primary MGTAB experiment.}
\label{tab:supp-mgtab-repro-config}
\end{minipage}

\paragraph{Seeds, repetitions, and uncertainty.}
Unless a table states otherwise, each backbone/prior configuration is one locked run with seed 42; a mean over 13 backbones is therefore an architecture average, not a 13-seed average. Table~\ref{tab:mgtab-finalizer-seed-stability} changes only the finalizer seed over five values while holding all upstream caches fixed. The main-paper MGTAB confidence intervals and sign-flip tests treat the 13 backbones as paired units because their predictions share the same test nodes. FraudBench uses one split and one seed over six generator-specific graphs; its across-generator deviations are descriptive. TeleAntiFraud-28k uses one source-disjoint split and one locked run per backbone.

\paragraph{Computing environment.}
The reported local experiments were executed on Windows 11 with an Intel Core i7-14650HX CPU, 32 GB RAM, and an NVIDIA GeForce RTX 5070 Ti Laptop GPU with 12 GB VRAM. The principal graph environment used Python 3.10.16, PyTorch 2.8.0 development build with CUDA 12.8, PyTorch Geometric 2.6.1, NumPy 1.26.4, pandas 2.2.3, scikit-learn 1.6.1, XGBoost 3.1.0, LightGBM 4.6.0, and CatBoost 1.2.8. TabICL v2 is run from its separate released environment and exchanges only train/dev/test prior-score files with the graph pipeline.

\subsection{Artifact Availability}

Upon publication, we will release preprocessing, training, evaluation, and paired-test scripts; configuration manifests; and redistributable split indices or predictions. Original datasets remain under their source licenses.

\section{YelpZip LLM Evidence Prior}
\label{sec:supp-yelpzip}

For YelpZip, the routed prior generator is instantiated by an LLM over compact evidence chains containing review text, rating, product-month context, reviewer history, burst behavior, and graph-neighbor summaries. These chains are consumed only by the router and LLM-prior branch for routed hard reviews; GNNs still receive only data-side features. The guarded prompt excludes labels, splits, GNN votes, and hard-region annotations. We use Qwen-Plus \cite{alibaba2026qwenplus} and DeepSeek-V4-Pro \cite{deepseek2026v4pro} prior files over 608,458 review IDs.

\paragraph{System prompt (compact guarded).}
\begingroup
\raggedright
You are a calibrated Yelp fake-review auditor. Estimate fake probability only for listed target reviews from leakage-free evidence: text, rating, product-month context, user history, burst behavior, and graph-neighbor summaries. Never use labels, tags, splits, GNN votes, or hard-region annotations. High risk requires combined coordination, template/duplicate behavior, user burst/repetition, or low-information text with contextual support; single weak cues are insufficient. Map ordinary/weak/moderate/clear/near-certain evidence to 0.05--0.25/0.25--0.45/0.45--0.65/0.65--0.90/\(>0.90\). Return JSON with continuous \texttt{p\_fraud} and evidence codes; avoid repeated values.
\par
\endgroup
\section{Multimodal FraudBench Evaluation}
\label{sec:supp-fraudbench}

\subsection{Dataset and Protocol}

FraudBench \cite{yan2026fraudbench} covers 29 product/service categories and contains 802 real-damaged claims and 988 fake-damaged claims from each of six image generators. Because it has no official graph, we build one graph per generator by linking claims that share a released user, item, or location identity. Joint entity grouping yields 1,056/363/371 train/development/test nodes with no cross-split entity edges. Separate generator graphs avoid connecting multiple edits of the same source image.

\subsection{Feature Isolation and Pipeline Components}

\paragraph{Graph branch.}
The GNNs receive only log user, item, and location degrees and the constructed graph. Images, text, ratings, platform/category fields, product metadata, and generator identity are excluded.

\paragraph{Multimodal evidence branch.}
The 238-dimensional evidence vector combines a 96-dimensional frozen EfficientNet-B0 image embedding \cite{tan2019efficientnet}, 14 decoded-pixel statistics, and 128 review-text dimensions. Learned transforms use training data only. Decision-related fields and file-level shortcuts, including ratings, dimensions, file size, and EXIF counts, are excluded.

\subsection{Results}

The backbone-only control is the first-stage GCN used by \fullmethod{}; the GNN ensemble combines the five downstream discovery models. Probability fusion combines the backbone and routed prior, while stacking w/o VN uses the backbone, prior, router, and route mask without structural injection. Table~\ref{tab:fraudbench-main} reports means over six generator-specific graphs.

\noindent\begin{minipage}{\linewidth}
\centering
\small
\setlength{\tabcolsep}{3.4pt}
\begin{tabular}{@{}lccc@{}}
\toprule
Method & Acc. & F1$_+$ & AUC \\
\midrule
Backbone only & 0.8275 & 0.8192 & 0.8438 \\
GNN ensemble & 0.8275 & 0.8192 & 0.8440 \\
Prob. fusion & 0.9533 & 0.9585 & 0.9870 \\
Stacking w/o VN & 0.9573 & 0.9620 & 0.9963 \\
\fullmethod{} & \textbf{0.9780} & \textbf{0.9800} & \textbf{0.9974} \\
\bottomrule
\end{tabular}
\normalsize
\captionof{table}{FraudBench macro means over six generator-specific graphs.}
\label{tab:fraudbench-main}
\end{minipage}

\noindent\begin{minipage}{\linewidth}
\centering
\small
\setlength{\tabcolsep}{3.0pt}
\begin{tabular}{@{}lccc@{}}
\toprule
Generator & Acc. & F1$_+$ & AUC \\
\midrule
GPT Image 2 & 0.9811 & 0.9828 & 0.9950 \\
Grok Imagine & 0.9838 & 0.9855 & 0.9987 \\
Nano Banana 2 & 0.9596 & 0.9624 & 0.9951 \\
Qwen Image 2.0 Pro & 0.9704 & 0.9737 & 0.9983 \\
Qwen Image Edit Max & 0.9865 & 0.9880 & 0.9981 \\
Wan 2.7 Image Pro & 0.9865 & 0.9878 & 0.9994 \\
\bottomrule
\end{tabular}
\normalsize
\captionof{table}{Per-generator \fullmethod{} test results on FraudBench.}
\label{tab:fraudbench-generator-results}
\end{minipage}

\fullmethod{} improves accuracy and F1$_+$ over no-VN stacking on five of six generators and AUC on three. The six graphs share the real claims, so they are not independent replications.

\subsection{Modality and Shortcut Audit}

Table~\ref{tab:fraudbench-modality-audit} audits retained evidence after removing decision and file-level shortcuts. Text supplies most ranking signal, while image features improve all three metrics.

\noindent\begin{minipage}{\linewidth}
\centering
\small
\setlength{\tabcolsep}{3.6pt}
\begin{tabular}{@{}lccc@{}}
\toprule
Evidence input & Acc. & F1$_+$ & AUC \\
\midrule
Visual embedding & 0.7633 & 0.8093 & 0.8570 \\
Decoded-pixel statistics & 0.5678 & 0.7081 & 0.5554 \\
All image features & 0.7579 & 0.8106 & 0.8589 \\
Text & 0.9218 & 0.9258 & 0.9934 \\
Image + text & 0.9623 & 0.9644 & 0.9971 \\
\bottomrule
\end{tabular}
\normalsize
\captionof{table}{FraudBench modality audit after removing decision and file-level shortcuts.}
\label{tab:fraudbench-modality-audit}
\end{minipage}

Generated reviews remain highly separable from real complaints, so the near-saturated AUC reflects dataset construction rather than broad multimodal generalization. Results use one split and seed, and the six generator graphs share real claims; the small AUC gain over stacking is therefore not statistically established.
\section{Audio--Text TeleAntiFraud-28k Evaluation}
\label{sec:supp-teleantifraud}

\subsection{Dataset and Feature Isolation}

TeleAntiFraud-28k \cite{ma2025teleantifraud} is evaluated under source-disjoint splits.

The legacy release yields 15,393 canonical calls with valid audio after invalid and duplicate records are quarantined. Complete source groups define 9,095/2,387/3,911 train/development/test calls. Each call forms a disconnected dialogue graph with one target call node, utterance nodes, bidirectional call--utterance edges, and consecutive-utterance edges. The resulting graphs contain 158,201 nodes and 540,446 directed edges, with no cross-split edges.

\paragraph{Graph branch.}
GNNs receive 19 node-type, speaker-role, turn-position, duration, pause, and call-level timing features. Transcripts, audio embeddings, source identities, collection metadata, and labels are excluded.

\paragraph{Evidence branch.}
Frozen XLSR-53 audio representations \cite{conneau2021xlsr} are window-pooled and reduced to 96 dimensions; multilingual MiniLM utterance representations \cite{wang2020minilm,reimers2019sentencebert} are call-pooled and reduced to 128 dimensions. Preprocessing is fitted on training call nodes only, and auxiliary utterance nodes receive zero evidence features.

\subsection{Results}

Table~\ref{tab:teleantifraud-main} reports macro means over 13 matched target backbones

\noindent\begin{minipage}{\linewidth}
\centering
\small
\setlength{\tabcolsep}{3.4pt}
\begin{tabular}{@{}lccc@{}}
\toprule
Method & Acc. & F1$_+$ & AUC \\
\midrule
Backbone only & 0.5806 & 0.7173 & 0.9000 \\
GNN ensemble & 0.5763 & 0.7151 & 0.9496 \\
Logit fusion & 0.8995 & 0.9141 & 0.9926 \\
Stacking w/o VN & 0.8841 & 0.9026 & 0.9935 \\
\fullmethod{} & \textbf{0.9353} & \textbf{0.9425} & \textbf{0.9936} \\
\bottomrule
\end{tabular}
\normalsize
\captionof{table}{TeleAntiFraud-28k results under source-disjoint splits. Matched methods are macro means over 13 target backbones.}
\label{tab:teleantifraud-main}
\end{minipage}

\noindent\begin{minipage}{\linewidth}
\centering
\small
\setlength{\tabcolsep}{2.0pt}
\begin{tabular}{@{}lcccccc@{}}
\toprule
& \multicolumn{3}{c}{Backbone only} & \multicolumn{3}{c}{\fullmethod{}} \\
\cmidrule(lr){2-4}\cmidrule(l){5-7}
Backbone & Acc. & F1$_+$ & AUC & Acc. & F1$_+$ & AUC \\
\midrule
MLP & 0.6091 & 0.7311 & 0.9286 & \textbf{0.9368} & \textbf{0.9437} & \textbf{0.9953} \\
GCN & 0.5960 & 0.7247 & 0.9314 & \textbf{0.9361} & \textbf{0.9431} & \textbf{0.9948} \\
GAT & 0.5809 & 0.7174 & 0.9291 & \textbf{0.9315} & \textbf{0.9393} & \textbf{0.9925} \\
GraphSAGE & 0.5699 & 0.7122 & 0.9091 & \textbf{0.9335} & \textbf{0.9410} & \textbf{0.9944} \\
GCNII & 0.5794 & 0.7165 & 0.9379 & \textbf{0.9333} & \textbf{0.9407} & \textbf{0.9953} \\
APPNP & 0.5398 & 0.6982 & 0.6667 & \textbf{0.9547} & \textbf{0.9589} & \textbf{0.9827} \\
GPRGNN & 0.5710 & 0.7125 & 0.9499 & \textbf{0.9333} & \textbf{0.9407} & \textbf{0.9953} \\
MixHop & 0.5653 & 0.7099 & 0.9252 & \textbf{0.9330} & \textbf{0.9405} & \textbf{0.9950} \\
SIGN & 0.5952 & 0.7245 & 0.9430 & \textbf{0.9366} & \textbf{0.9435} & \textbf{0.9957} \\
ChebNet & 0.5796 & 0.7168 & 0.8975 & \textbf{0.9320} & \textbf{0.9397} & \textbf{0.9950} \\
GraphConv & 0.5761 & 0.7150 & 0.9562 & \textbf{0.9338} & \textbf{0.9412} & \textbf{0.9950} \\
TAGCN & 0.5866 & 0.7201 & 0.9202 & \textbf{0.9317} & \textbf{0.9395} & \textbf{0.9949} \\
GIN & 0.5988 & 0.7261 & 0.8044 & \textbf{0.9328} & \textbf{0.9403} & \textbf{0.9916} \\
\midrule
Mean & 0.5806 & 0.7173 & 0.9000 & \textbf{0.9353} & \textbf{0.9425} & \textbf{0.9936} \\
\bottomrule
\end{tabular}
\normalsize
\captionof{table}{TeleAntiFraud-28k matched results for all 13 target backbones.}
\label{tab:teleantifraud-backbones}
\end{minipage}

\fullmethod{} improves all three metrics for every target backbone in Table~\ref{tab:teleantifraud-backbones}.

\clearpage
\bibliography{aaai2027}
	
\end{document}